\documentclass{article}

\usepackage[main, final]{neurips_2026}

\usepackage[utf8]{inputenc} % allow utf-8 input
\usepackage[T1]{fontenc}    % use 8-bit T1 fonts
\usepackage{hyperref}       % hyperlinks
\usepackage{url}            % simple URL typesetting
\usepackage{booktabs}       % professional-quality tables
\usepackage{amsfonts}       % blackboard math symbols
\usepackage{nicefrac}       % compact symbols for 1/2, etc.
\usepackage{microtype}      % microtypography
\usepackage{xcolor}         % colors
\usepackage[dvipsnames]{xcolor}
\usepackage{graphicx}
\usepackage{amsmath}
\usepackage{multirow}
\usepackage{multicol}
\usepackage{booktabs}
\usepackage{cleveref}
\usepackage{makecell}
\usepackage{amssymb}
\usepackage{comment}
\usepackage{algorithm}
\usepackage{algpseudocode}
\usepackage{wrapfig}
\usepackage{caption}

\newcommand{\datasetfirst}{PolyMNIST}
\newcommand{\datasetsecond}{CelebAMask-HQ}
\newcommand{\datasetthird}{PIE-Bench++}
\newcommand{\datasetfourth}{ROCO}
\newcommand{\ourmethod}{SnB}
\title{Split and Bridge: Multimodal Generation via Diffusion Bridging}

\author{%
  Ahmad Arrabi$^{1}$\thanks{\texttt{\{ahmad.arrabi, safwan.wshah\}@uvm.edu}} \quad Xiaohan Zhang$^{1}$ \quad Xingyu Li$^{2}$ \quad Safwan Wshah$^{1\ast}$ \\[0.5em]
  $^{1}$Department of Computer Science, University of Vermont \\
  $^{2}$Center for Excellence in Brain Science and Intelligence Technology
}

\begin{document}

\maketitle
\begin{abstract}
Visual data are often expressed through multiple complementary modalities (e.g., images and segmentation masks), each capturing different aspects of the same underlying structure. Most existing multimodal methods adopt a top-down paradigm, learning a shared latent variable through joint training from scratch, which limits scalability. We propose Split and Bridge (\ourmethod{}), a bottom-up framework for multimodal generation that couples pretrained unimodal diffusion models at sampling time. Rather than learning a unified latent space, \ourmethod{} generates coherent multimodal samples by following a shared diffusion trajectory up to a split point and then branching into modality-specific trajectories guided by a bridge module. This design enables plug-and-play reuse of powerful pretrained models without joint training, making it scalable and easily adaptable. Across standard benchmarks (\datasetfirst{}, \datasetsecond{}) and more complex settings (\datasetthird{}, \datasetfourth{}), \ourmethod{}  achieves a strong balance between sample quality and coherence. We show that \ourmethod{} naturally extends to downstream zero-shot tasks such as interleaved co-generation, image editing, and style coupling/decoupling. Our results demonstrate that bottom-up generation provides a practical framework that complements existing top-down multimodal approaches. Code is available at \url{https://github.com/AhmadArrabi/SnB} 
\end{abstract}

%Visual data are naturally expressed through multiple complementary modalities (e.g., images and segmentation masks), each capturing distinct aspects of the same underlying structure. Most existing multimodal methods adopt a top-down paradigm, learning a shared latent variable through joint training from scratch, which limits scalability. We propose Split and Bridge (SnB), a bottom-up framework for multimodal generation that couples pretrained unimodal diffusion models at sampling time. Rather than learning a unified latent space, SnB generates coherent multimodal samples by following a shared diffusion trajectory up to a split point, after which modality-specific processes are guided via a bridge module. This design enables plug-and-play reuse of powerful pretrained models without joint training, making it scalable and easily adaptable. Across standard benchmarks (PolyMNIST, CelebAMask-HQ) and more complex settings (PIE-Bench++, ROCO), SnB consistently achieves a strong balance between sample quality and coherence, as we report a 33% increase in the coherence score on CelebAMask-HQ over prior top-down approaches. We show that SnB naturally extends to downstream zero-shot tasks such as interleaved co-generation, image editing, and style transfer. Our results demonstrate that bottom-up generation offers a practical and scalable alternative to top-down multimodal models.
\section{Introduction}
\label{sec:intro}
Different visual modalities capture multiple complementary aspects of the same underlying scene. For example, a segmentation mask highlights geometric shapes, whereas the image itself captures details about the scene. This variety of visual representations increases the demand for models that capture their shared underlying structure as well as modality-specific information. A particularly powerful formulation of this goal is multimodal generation, which aims to learn a joint distribution that enables the simultaneous sampling of coherent and high-fidelity samples of multiple modalities~\cite{mvrl,early1,elbo}. Accurate modeling of this joint distribution implies capturing the underlying mechanism that generated these modalities. This capability benefits multiple downstream applications such as representation learning~\cite{representation}, interleaved generation~\cite{interleaved1,interleaved2}, story telling~\cite{storytelling}, image editing, and medical imaging~\cite{medical1}.

Existing multimodal generation methods rely on a \textit{top-down} approach that learns a hidden latent variable $z$ that encompasses all modalities through joint training from scratch~\cite{mvrl, mmvae, mmvae_plus, elbo, dmvae, mmjsd}. However, empirical evidence reveals several critical limitations of top-down approaches: \textbf{1)} They fail to scale to complex, high-resolution visual data (e.g., $512\times512$ images) due to expensive latent bottleneck design (see \Cref{sec:learnable_bridge_exp}), \textbf{2)} They depend on \textit{restrictive} joint-posterior assumptions, such as the Product of Experts (PoE) or Mixture of Experts (MoE)~\cite{mvae,mmvae,elbo}, \textbf{3)} The unified latent design imposes rigidity, limiting adaptability to diverse downstream tasks (see \Cref{sec:scalability}), and \textbf{4)} They suffer from the coherence-quality trade-off, mainly in VAE-based formulations~\cite{limitations_vae} (see \Cref{sec:identity_bridge_exp}). Recent diffusion-based extensions (e.g., MM-Diffusion, MLD~\cite{mmdiffusion,mld}) partially alleviate fidelity issues. However, they still inherit key limitations as they rely on modality-specific architectures and require large scale training from scratch with paired data. For all those reasons, the dependence on top-down modeling for multimodal generation limits scalability in practice.

In contrast, a \textit{bottom-up} approach adopts a compositional design, where modality-specific models are developed independently and later coupled for cross-modal interactions. This paradigm offers greater flexibility and scalability by utilizing strong pretrained unimodal models. Similar organizational principles are observed in biological systems, where distinct sensory pathways process different modalities while interacting before higher-level integration~\cite{neuro1,neuro2,neuro3,neuro4,neuro5,neuro6}. 

Motivated by this perspective, we propose \textbf{S}plit a\textbf{n}d \textbf{B}ridge (\ourmethod{}), a bottom-up framework for multimodal generation. \ourmethod{} leverages pretrained unimodal diffusion models (e.g., LDM~\cite{ldm}, SDXL~\cite{flux}, and FLUX~\cite{flux}) and introduces a bridge module that connects modalities during sampling. This formulation constructs the joint generative process through coupling modality-specific information. As a result, it relaxes restrictive joint-posterior assumptions while retaining the scalability of modern diffusion models. Given two (or more) diffusion models, we share a common reverse trajectory up to a split time step, then diverge into modality-specific branches via a bridge module. Since all trajectories originate from a common intermediate state, we hypothesize that the resulting samples share structure while preserving modality-specific information. \ourmethod{} is a general framework, and we propose multiple bridge designs based on dataset complexity and the required task. This modular design makes \ourmethod{} flexible and easier to scale while gaining competitive performance in multimodal generation.

To illustrate \ourmethod{}'s flexibility and performance, we evaluate on standard multimodal generation benchmarks, including \datasetfirst{}~\cite{elbo} and \datasetsecond{}~\cite{celeb}. \ourmethod{} consistently produces coherent multimodal samples while maintaining high generation fidelity. \ourmethod{} supports both unconditional joint generation and conditional generation within a unified framework while being competitive with top-down approaches. Beyond benchmark evaluation, we further demonstrate the practical flexibility of \ourmethod{} through downstream tasks on \datasetthird{}~\cite{pie_bench}, including interleaved co-generation and image editing. We also apply \ourmethod{} on a practical use case, multimodal MRI generation using the \datasetfourth{}~\cite{roco} radiology dataset. In summary, our contributions are as follows:%\vspace{-5pt}
\begin{itemize}
    \item We propose Split and Bridge (\ourmethod{}), a bottom-up framework for multimodal generation that composes pretrained unimodal diffusion models without requiring joint retraining. \ourmethod{} avoids the scalability limitations and restrictive assumptions of top-down approaches.%\vspace{-2pt}
    \item We introduce bridge mechanisms that enable controlled coupling of modality trajectories, supporting different levels of modality interaction.%\vspace{-2pt}
    %\item We demonstrate that \ourmethod{} achieves superior performance over top-down multimodal methods in both sample quality and coherence.\vspace{-2pt}
    \item We demonstrate that \ourmethod{} achieves superior performance compared to top-down multimodal methods in both sample quality and coherence on \datasetsecond{}, while providing competitive performance on \datasetfirst{}.
    \item We show that \ourmethod{} can be extended to downstream tasks, including interleaved co-generation, image editing, and MRI synthesis, highlighting its flexibility and practical applicability.
\end{itemize}
\section{Related Work}
\label{sec:related}
\paragraph{VAE-based methods}
Multimodal VAE methods were initially introduced to address the multimodal generation task through a top-down hidden variable approach~\cite {early1,mvae,vae_google}. Early works extended the VAE framework by training multiple encoders, one for each modality, to infer a shared latent variable $\mathbf{z}$~\cite{early1}. However, these models neither explored the structure of the latent representation nor scaled effectively as the number of modalities increased~\cite{mmvae_plus}. To tackle this limitation, more theoretically grounded approaches~\cite{mvae,mmvae} have been proposed to explicitly model the multimodal posterior $p(\mathbf{z} | \textbf{x}_1, \dots,\textbf{x}_m)$, where $m$ is the number of modalities. These works introduced different assumptions on the joint posterior, such as the Product of Experts (PoE)~\cite{mvae}, Mixture of Experts (MoE)~\cite{mmvae}, and Mixture of Product of Experts (MoPoE)~\cite{elbo}, leading to enhanced modeling. Further refinements addressed specific weaknesses of multimodal VAEs like proposing improved training objectives~\cite{mmjsd,mvrl,mmvae_plus,vae_diffusion,sbmvae,mmvmvae} and learning disentangled representations to separate modality-private and -shared information~\cite{dmvae,mmvae_plus}. Recent work in \cite{shala} has shown promising multimodal generation capabilities, but still faces top-down limitations like controllability and flexibility issues.

\paragraph{Diffusion-based methods}
Recent advances in diffusion models have inspired many works to apply them in multimodal problems. MM-Diffusion~\cite{mmdiffusion} introduced the first diffusion-based model for joint audio–video generation. It assumes two independent forward processes per modality and models their correlation through a single shared reverse process implemented via a U-Net with multimodal blocks. Although MM-Diffusion demonstrates impressive results in the audio–video domain, its architecture is highly specialized and tailored to this specific task, limiting its generalization to visual modalities. Similarly, MLD~\cite{mld} proposed a diffusion-based joint generation framework, but follows a top-down paradigm in which the diffusion process is applied on a concatenation of all latent representations encoded through pretrained encoders. This design leads to an expensive multi-stage training procedure that restricts scalability. Another family of diffusion models are diffusion bridges, which learn a stochastic transport process between modalities, mapping samples from one domain to another~\cite{bridge_1,bridge_2,bridge_3}. However, a key difference between multimodal models and diffusion bridges is that the latter are \textit{not} designed for joint generation. Instead, they address a transport problem between existing modalities, typically assuming a predefined source-target relationship.

%\vspace{-10pt}
\paragraph{Connection to Conditional Generation}
Multimodal generation has advanced significantly in conditional generation, where models learn a directed mapping from a source modality to a target~\cite{controlnet,controlnet++,t2i,controlnet3}. Although effective for tasks like text-to-image~\cite{controlnet3} or image-to-image translation~\cite{sketch2anim}, these methods rely on strong supervision (e.g., prior image) and thus \textit{fall outside the scope of joint generation}. More recently, large foundation diffusion models such as CoDI and CoDI-2~\cite{codi,codiv2} have enabled any-to-any generation across modalities. However, they remain inherently conditional, rather than jointly sampling from a true multimodal distribution (co-generation without priors).

%In contrast, multimodal generation aims to enable both co-generation and conditional generation across modalities under weak supervision. This encourages models to learn more expressive joint distributions that enhance representation learning~\cite{mmvae,mvae}. Despite advancements in multimodal VAEs, they still require explicit modeling of a shared latent bottleneck, often with strong assumptions on the posterior form (e.g., PoE, MoE), resulting in the top-down paradigm. These assumptions can lead to misrepresenting the true data distribution~\cite{limitations_vae}. Furthermore, as shown in \Cref{sec:experiments}, their scalability to high-resolution images remains limited. To tackle this limitation, we introduce a diffusion-based alternative that supports high-resolution multimodal generation. \ourmethod{} adopts a bottom-up formulation that leverages pretrained unimodal diffusion models enabling coherent multimodal generation.

%\vspace{-5pt}
\section{Methodology}
\label{sec:methods}
\begin{figure}
    \centering
    %\vspace{-1pt}
    \includegraphics[width=\textwidth,trim={0.5cm 0cm 0.5cm 2.5cm},clip,page=9]{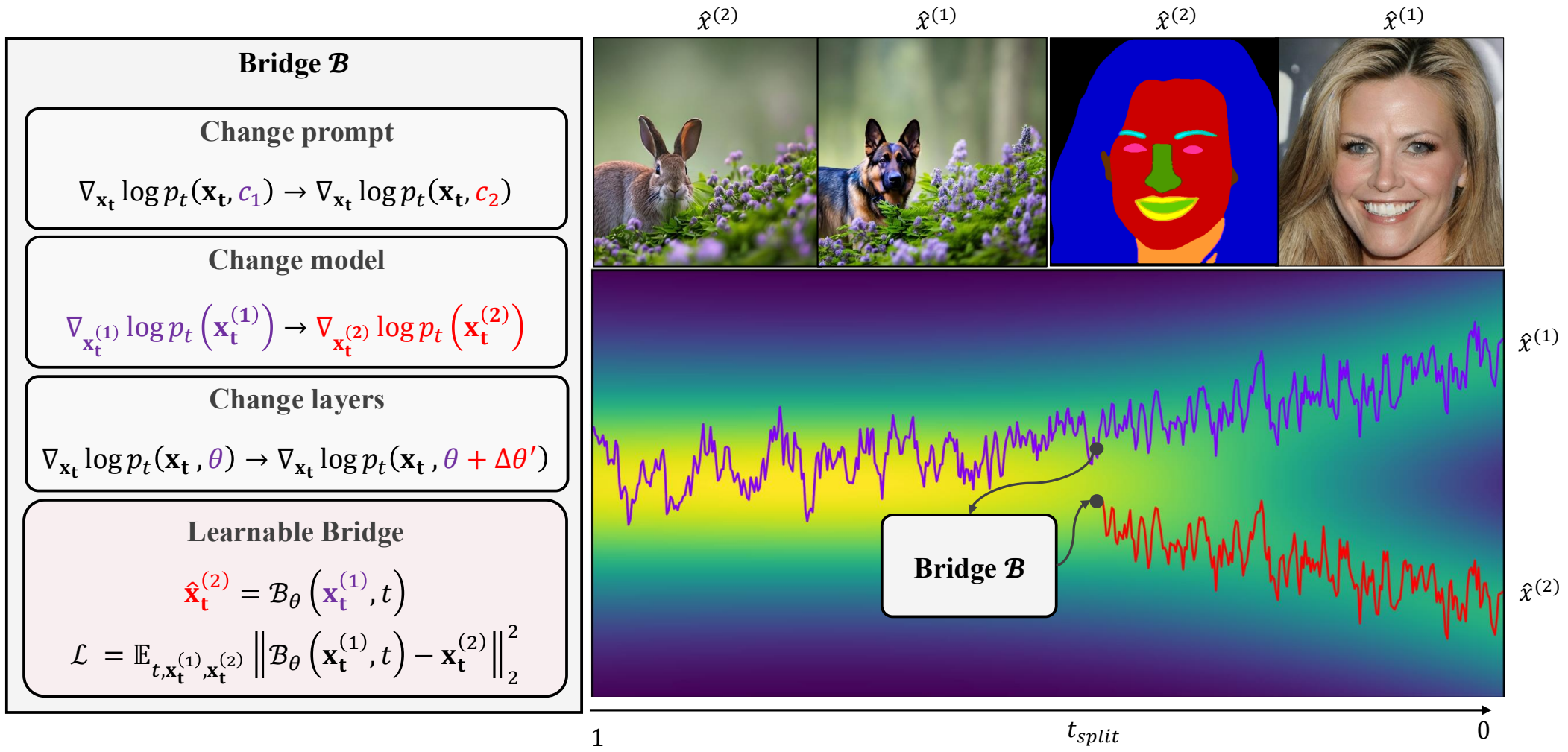}
    \caption{\ourmethod{} leverages pretrained diffusion models by coupling reverse trajectories during sampling. We start from a common state sampled from a prior, then at $t_{split}$, a bridge module is used to implement a score intervention on the reverse process. This intervention can be through modifying conditioning signals (e.g., prompts), changing diffusion models that estimate the score functions, or changing certain layers in the diffusion models. This flexibility applies \ourmethod{} in various tasks, like interleaved co-generation (top left pair) and multimodal generation (top right pair). Learnable bridges allows stronger trajectory manipulation in mapping one modality to another.}
    \label{fig:main_fig}
    \vspace{-10pt}
\end{figure}
\paragraph{Diffusion Models Preliminaries}
Diffusion models learn a data distribution $p_{\text{data}}(\mathbf{x})$ through a continuous-time stochastic process that progressively perturbs data into noise and reverses this process for generation~\cite{song_score-based_2021}. The forward process is defined by the stochastic differential equation (SDE),
\begin{equation}
  \label{eq:SDE_forward}
  d\mathbf{x}_t = f(\mathbf{x}_t, t)\,dt + g(t)\,d\mathbf{w}_t
\end{equation}
where $f(\mathbf{x}_t, t) \colon \mathbb{R}^d \times [0,1] \to \mathbb{R}^d$ is the drift coefficient, $g(t) \colon [0,1] \to \mathbb{R}$ is the scalar diffusion coefficient, and $\mathbf{w}_t \in \mathbb{R}^d$ is a standard Wiener process. The process is defined over $t \in [0,1]$, where $\mathbf{x}_t$ evolves from clean data $\mathbf{x}_0 \sim p_{\text{data}}(\mathbf{x})$, and as $t \to 1$, $\mathbf{x}_t$ converges to a tractable Gaussian prior $p_1(\mathbf{x}_1) \approx \mathcal{N}(\mathbf{0}, \mathbf{I})$.
\cite{song_score-based_2021} and \cite{SDE_reverse} showed that \Cref{eq:SDE_forward} admits a corresponding reverse-time SDE that reconstructs data by inverting the noising dynamics,
\begin{equation}
  \label{eq:SDE_reverse}
  d\mathbf{x}_t
    = \Bigl[f(\mathbf{x}_t, t) - g(t)^2\,\nabla_{\mathbf{x}_t} \log p_t(\mathbf{x}_t)\Bigr]dt
    + g(t)\,d\bar{\mathbf{w}}_t
\end{equation}
where $p_t(\mathbf{x}_t)$ is the marginal density of $\mathbf{x}_t$ under the forward process, and $\bar{\mathbf{w}}_t$ is a standard Wiener process running in reverse time (i.e., as $t$ decreases from $1$ to $0$). The term $\nabla_{\mathbf{x}_t} \log p_t(\mathbf{x}_t)$ is the Stein score function of the marginal distribution at time $t$. By approximating it, one can simulate the reverse-time SDE starting from $\mathbf{x}_1 \sim \mathcal{N}(\mathbf{0}, \mathbf{I})$ to generate samples approximately distributed according to $p_{\text{data}}(\mathbf{x})$.

\paragraph{Problem Formulation} Let $\mathbf{X} = \{\mathbf{x}^{(1)}, \dots, \mathbf{x}^{(m)}\}$ denote a tuple of $m$ modalities drawn from an unknown joint distribution $p_{\text{data}}(\mathbf{x}^{(1)}, \dots, \mathbf{x}^{(m)})$. The goal of multimodal generation is to generate samples $\mathbf{\hat{X}}$ that show unimodal fidelity and cross-modal coherence. Here, coherence means that all generated modalities share semantic factors (e.g., object geometry, facial attributes, or digit identity).

Multimodal generation is often tackled from a top-down approach in learning a hidden variable $\mathbf{z}$ by approximating its posterior in the following general form~\cite{elbo},
\begin{equation}
q_\phi(\mathbf{z} \mid \mathbf{x}^{(1)}, \cdots, \mathbf{x}^{(m)}) = 
    \mathcal{C}\left(q_{\phi_1}(\mathbf{z} \mid \mathbf{x}^{(1)}), \cdots, q_{\phi_m}(\mathbf{z} \mid \mathbf{x}^{(m)}) \right)
\end{equation}
where $\mathcal{C}$ is an aggregation function that combines all individual posteriors per modality, $\phi=\{\phi\}_{j=1}^m$ is the model's parameters. For example, in~\cite{mvae} $\mathcal{C}$ was defined as a product function, which led to the PoE formulation, ~\cite{mmvae} used a summation, leading to the MoE formulation. 
\begin{wrapfigure}{r}{0.55\textwidth}
%\vspace{-5pt} % Adjust to align top of text
\begin{minipage}{\linewidth}
\begin{algorithm}[H]
\caption{\ourmethod{}}
\label{alg:bridge_sampling}
\begin{algorithmic}[1]
\State \textbf{Input:} Diffusion models $\mathcal{M}_1, \mathcal{M}_2$, bridge $\mathcal{B}$, $t_{\text{split}}$
\State Sample $\mathbf{x}_1^{(1)} \sim \mathcal{N}(\mathbf{0}, \mathbf{I})$
\For{$t = 1 \to 0$}
    \If{$t = t_{\text{split}}$} \Comment{\textcolor{RoyalBlue}{split}}
        \State $\mathbf{x}_t^{(2)} \gets \mathcal{B}(\mathbf{x}_t^{(1)}, t)$ \Comment{\textcolor{RoyalBlue}{bridge}}
    \EndIf
    \State $\hat{\mathbf{y}}_t^{(1)} \gets \mathcal{M}_1(\mathbf{x}_t^{(1)}, t)$ \Comment{\textcolor{RoyalBlue}{noise / velocity / score}}
    \State $\mathbf{x}_{t-\Delta t}^{(1)} \gets \mathrm{Denoise}(\mathbf{x}_t^{(1)}, \hat{\mathbf{y}}_t^{(1)})$
    \If{$t \le t_{\text{split}}$}
        \State $\hat{\mathbf{y}}_t^{(2)} \gets \mathcal{M}_2(\mathbf{x}_t^{(2)}, t)$
        \State $\mathbf{x}_{t-\Delta t}^{(2)} \gets \mathrm{Denoise}(\mathbf{x}_t^{(2)}, \hat{\mathbf{y}}_t^{(2)})$
    \EndIf
\EndFor
\State \textbf{Output:} $\{\mathbf{x}_0^{(1)}, \, \mathbf{x}_0^{(2)}\}$
\end{algorithmic}
%\begin{algorithmic}[1]
%\State \textbf{Input:} Diffusion models $\mathcal{M}_1, \mathcal{M}_2$, bridge $\mathcal{B}$, $t_{split}$
%\State Sample $\mathbf{x}_1^{(1)} \sim \mathcal{N}(0,I)$
%\For{$t = 1 \to 0$}
%    \State $\hat{\mathbf{y}}_t^{(1)} \gets \mathcal{M}_1(\mathbf{x}_t^{(1)}, t)$ 
%    \Comment{\textcolor{RoyalBlue}{noise / velocity / score}}
%    \State $\mathbf{x}_{t-\Delta t}^{(1)} \gets \mathrm{Denoise}(\mathbf{x}_t^{(1)}, \hat{\mathbf{y}}_t^{(1)})$
%    \If{$t = t_{split}$}
%        \Comment{\textcolor{RoyalBlue}{split}}
%        %\State Duplicate $x_t^{(1)}$ to obtain $\{x_t^{(1)}, x_t^{(1)}\}$
%        \State $\hat{\mathbf{x}}_t^{(2)} \gets \mathcal{B}(\mathbf{x}_t^{(1)}, t)$
%        \Comment{\textcolor{RoyalBlue}{bridge}}
%        %\State Denoising $\mathbf{x}_t^{(2)}$ with $\mathcal{M}_2$ until $t = 0$
%    \EndIf
%    \If{$t < t_{split}$}
%        \State $\hat{\mathbf{y}}_t^{(2)} \gets \mathcal{M}_2(\mathbf{x}_t^{(2)}, t)$ 
%        %\Comment{\textcolor{RoyalBlue}{noise / velocity / score}}
%        \State $\mathbf{x}_{t-\Delta t}^{(2)} \gets \mathrm{Denoise}(\mathbf{x}_t^{(2)}, \hat{\mathbf{y}}_t^{(2)})$ 
%    \EndIf
%\EndFor
%\State \textbf{Output:} $\{\mathbf{x}_0^{(1)},\, \mathbf{x}_0^{(2)}\}$
%\end{algorithmic}
\end{algorithm}
\end{minipage}
%\vspace{-10pt} % Adjust to align bottom of text
\end{wrapfigure}
\vspace{-12pt}
\paragraph{Split and Bridge (SnB)} We don't limit multimodal generation to restrictions over the posterior's functional form. We consider the problem alternatively from a \textit{bottom-up} approach. Rather than explicitly learning posteriors, or training large multi-branch models~\cite{mld,codi,mmdiffusion}, we leverage pretrained unimodal diffusion models, and learn to \textit{bridge} their trajectories \textit{during sampling} to achieve joint multimodal generation. We consider each modality modeled by its reverse SDE following \Cref{eq:SDE_reverse}, with a drift learned through training its individual diffusion model. 

Multimodal generation requires joint sampling of a modality set $\mathbf{X}$. Thus, we propose a new sampling framework, as illustrated in~\Cref{fig:main_fig}, where we first sample a single diffusion trajectory (any chosen initial modality) backward in time starting at \(t=1\) to an intermediate step \(t_{split}\). Then, we divide the denoising process into $m$ independent processes, one per modality. We hypothesize that this split generates a latent noisy representation that contains the initial structure of the synthesized object (e.g., a noisy face of a person), which incentivizes correlated trajectories. At $t_{split}$, we introduce a bridge module $\mathcal{B}:\mathbb{R}^d\rightarrow \mathbb{R}^k$ that transports one modality closer to the other through a transport operation, where $d$ and $k$ are the dimensions of the corresponding modalities. After $t_{split}$, each trajectory would be determined by its own reverse process that is driven by its modality-specific diffusion model. More formally, we can define the joint sampling process as multiple reverse SDEs as shown in~\Cref{eq:split_sampling_SDE}. For clarity and notation simplicity, we present the case of $m=2$ but the formulation extends to $m$ modalities.

\begin{equation}
\label{eq:split_sampling_SDE}
\begin{split}
    d\mathbf{x}_t^{(1)} &= \Big[f_1(\mathbf{x}_t^{(1)}, t) - g_1(t)^2 \nabla_{\mathbf{x}_t^{(1)}} \log p_t(\mathbf{x}_t^{(1)})\Big]\,dt + g_1(t)d\mathbf{w}_t \\
    d\mathbf{x}_t^{(2)} &=
    \begin{cases}
         d\mathbf{x}_t^{(1)}                         &: t > t_{split} \\
         \mathcal{B}(\mathbf{x}_t^{(1)}, t)          &: t = t_{split} \\
         \Big[f_2(\mathbf{x}_t^{(2)}, t) - g_2(t)^2 \nabla_{\mathbf{x}_t^{(2)}} \log p_t(\mathbf{x}_t^{(2)})\Big]\,dt + g_2(t)d\mathbf{w}_t     &: t < t_{split}
    \end{cases}
\end{split}
\end{equation}
\Cref{alg:bridge_sampling} implements \ourmethod{} in ~\Cref{eq:split_sampling_SDE}, where $\mathcal{M}_1 \text{ and } \mathcal{M}_2$ are diffusion models pretrained for each modality. \ourmethod{} is agnostic to the diffusion parameterization (e.g., score-based~\cite{song_score-based_2021}, noise-prediction~\cite{ddpm}, or velocity-prediction~\cite{velocity}). Sampling begins by drawing $\mathbf{x}_1^{(1)} \sim \mathcal{N}(\mathbf{0}, \mathbf{I})$ and denoising it along the reverse trajectory defined by $\mathcal{M}_1$ until the split timestep $t_{split}$. At $t_{split}$, a second trajectory is initialized via the bridge module, $\hat{\mathbf{x}}_{t_{split}}^{(2)} = \mathcal{B}\big(\mathbf{x}_{t_{split}}^{(1)}, t_{split}\big),$ which maps the state from the first modality to the second. Finally, the two representations $\mathbf{x}_t^{(1)}$ and $\mathbf{x}_t^{(2)}$ are denoised in parallel by their respective models to produce the coherent output set $\{\mathbf{x}_0^{(1)}, \mathbf{x}_0^{(2)}\}$. 
%Generation starts by sampling $\mathbf{x}_T^{(1)}$ from the prior and denoising it via $\mathcal{M}_1$ reaching the split timestep $t_{split}$. At this point, we generate a second trajectory $\hat{\mathbf{x}}_t^{(2)}$ using the bridge module $\mathcal{B}$ which transports $\mathbf{x}_t^{(1)}$ to the other modality space. 

%\vspace{-6pt}
\paragraph{Bridge Module $\mathcal{B}$} The transport strength needed at $t_{split}$ depends largely on the discrepancy between modalities and the complexity of the generated data. We therefore introduce different bridges that can be applied to different data complexity levels and tasks. This flexibility is mainly a result of the bottom-up formulation of \ourmethod{}, as it allows leveraging strong modality-specific models.

\noindent\textbf{i. Identity Bridge}
We first define $\mathcal{B}$ as an identity mapping as shown in~\Cref{eq:bridge_identity}. Despite its simplicity, the identity bridge enables multimodal generation by switching the score function that governs the reverse process after $t_{split}$. Concretely, a shared latent state is continued under a new score function, effectively transitioning from one data distribution to another without explicit transport: $
\nabla_{\mathbf{x}^{(1)}_t} \log p_t(\mathbf{x}^{(1)}_t) \rightarrow \nabla_{\mathbf{x}^{(2)}_t} \log p_t(\mathbf{x}^{(2)}_t)$. This perspective shows that multimodal generation can be achieved through interventions on the score function during sampling, rather than via joint training. The identity bridge therefore represents the minimal mechanism for coupling modalities.
\begin{equation}
\label{eq:bridge_identity}
\hat{\mathbf{x}}_{t_{split}}^{(2)} = \mathcal{B}(\mathbf{x}_{t_{split}}^{(1)}, t_{split}) = \mathbf{x}_{t_{split}}^{(1)}
\end{equation}
We consider several ways in which a score intervention can happen, all unified under the same principle:
1) \textit{Model switch}, where the reverse process transitions between pretrained diffusion models corresponding to different modalities. This case is effective in simple data structures where a model per modality is feasible.
2) \textit{Condition switch}, where the score is modified via conditioning signals (e.g., prompt changes in text-guided diffusion models), enabling controlled semantic variation, and 
3) \textit{Parameter switch}, where lightweight adaptations (e.g., LoRA\cite{lora}) alter the score function without replacing the full model allowing zero-shot compositional generation (see \Cref{app:downstream}).

Different kinds of score interventions allow \ourmethod{} to integrate with pretrained and foundation diffusion models. For instance, in text-conditioned models, a simple prompt change at $t_{split}$ acts as an effective bridge, preserving global structure while inducing targeted variations aligned with the new condition. The identity bridge is most effective when modalities share similarity, such that the intermediate state provides a meaningful initialization across branches. In larger modality gaps, stronger bridge designs that explicitly transform the latent state would be needed (see \Cref{sec:learnable_bridge_exp}).

\noindent\textbf{ii. Learnable Bridge}
When the domain gap between modalities is large (e.g., faces and segmentation maps), directly switching the drift from $f_1$ to $f_2$ at $t_{split}$ leads to poor generation (as will be shown in \Cref{sec:ablation}). The intermediate sample $\mathbf{x}_{t_{split}}^{(1)}$ results in an improper initialization for the new trajectory. To address this, we introduce a learnable bridge network $\mathcal{B}_\theta$ that maps samples across modalities at the splitting point. The bridge acts as a drift correction mechanism, projecting $\mathbf{x}_{t_{split}}^{(1)}$ onto a suitable region of the target modality's latent space. Formally, $\mathcal{B}_\theta$ is defined as follows,
\begin{equation}
    \hat{\mathbf{x}}_{t_{split}}^{(2)} = \mathcal{B}_\theta(\mathbf{x}_{t_{split}}^{(1)}, t_{split})
\end{equation}
After bridging, each modality continues its own reverse diffusion trajectory driven by its diffusion model that controls the drift $f_i$, where $i$ is the modality index. Because the transition occurs under shared noise, the generated modalities remain stochastically aligned. We implement $\mathcal{B}_\theta$ as a time-conditioned U-Net that takes a noisy sample $\mathbf{x}_{t_{split}}^{(1)}$ as input and outputs its aligned counterpart $\mathbf{x}_{t_{split}}^{(2)}$. During training, we sample time steps $t_{split} \sim \mathcal{U}(0, 1)$ uniformly to ensure $\mathcal{B}_\theta$ can operate anywhere on the diffusion trajectory $[0,1]$. The bridge is trained using the following reconstruction loss,
\begin{equation}
\label{eq:mse_loss}
    \mathcal{L}_{\text{bridge}} 
    = \mathbb{E}_{t, \mathbf{x}_t^{(1)}, \mathbf{x}_t^{(2)}} \| \mathcal{B}_\theta(\mathbf{x}_t^{(1)}, t) - \mathbf{x}_t^{(2)} \|_2^2
\end{equation}
A Mean Squared Error (MSE) objective encourages $\mathcal{B}_\theta$ to approximate the conditional expectation, which can reduce stochastic variation in $\hat{\mathbf{x}}_t^{(2)}$. As a result, the predicted state may become overly smooth and misaligned with the noise level expected by the diffusion process at $t_{split}$. Inspired by~\cite{meng_sdedit_2022}, we reintroduce noise into $\hat{\mathbf{x}}_t^{(2)}$ using the diffusion scheduler (derived from the coefficient $g_2(t)$ in~\Cref{eq:SDE_forward}) to ensure consistency with the reverse trajectory. We provide additional details and a formal derivation in \Cref{app:noise_injector}.
\begin{wrapfigure}{r}{0.5\textwidth}
    \centering
    \vspace{15pt}
    \includegraphics[width=\linewidth]{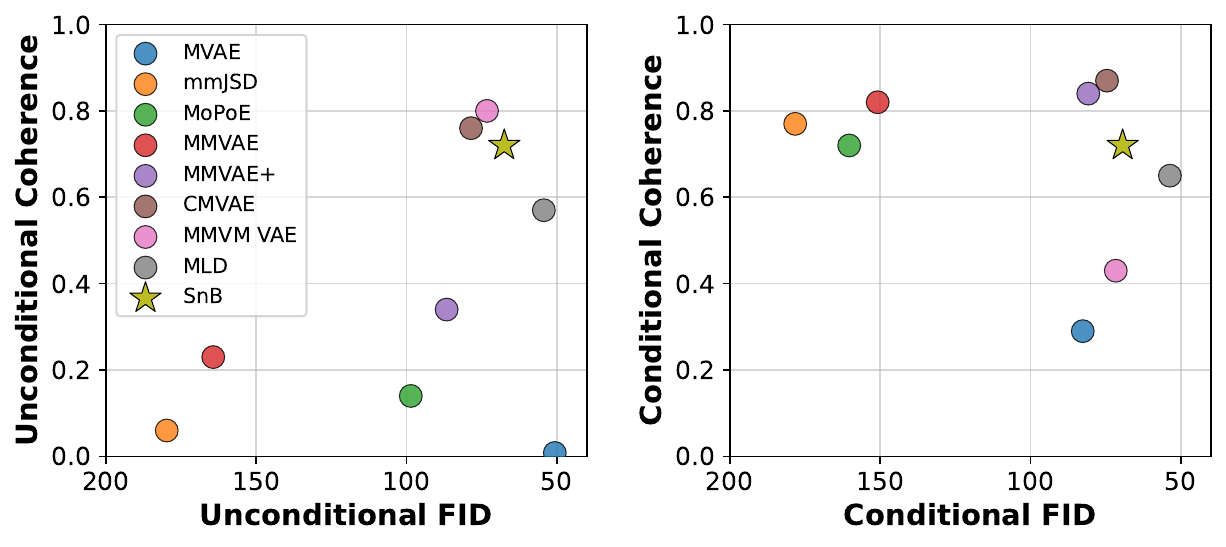}
    \caption{PolyMNIST results using the identity bridge with \ourmethod{}. \ourmethod{} strikes a balance between coherence and quality across the 5 modalities.}
    \vspace{-20pt}
    \label{fig:results_PolyMNIST}
\end{wrapfigure}
\vspace{-15pt}
\section{Experiments}
\vspace{-5pt}
\label{sec:experiments}
\paragraph{Implementation Details}
We evaluate \ourmethod{} on \datasetfirst{}~\cite{elbo}, \datasetsecond{}~\cite{celeb}, \datasetthird{}~\cite{pie_bench}, and \datasetfourth{}~\cite{roco}. We compare against top-down baselines, including VAE-based methods (MVAE~\cite{mvae}, MMVAE, MMVAE+, MoPoE, mmJSD~\cite{mmjsd}, MMVM VAE\cite{mmvmvae}) and diffusion-based approaches (MLD\cite{mld} and MM-Diffusion variants with both Stable Diffusion~\cite{ldm} and ResNet~\cite{resnet} U-Net backbones for a fair comparison). Image quality is measured using FID, and segmentation quality using F1 score when applicable. For coherence, we follow prior work on \datasetfirst{} using pretrained classifiers~\cite{elbo}. We also introduce a learned metric for \datasetsecond{} that measures the cosine similarity between $\ell_2$-normalized embeddings of generated modalities. The full implementation details are provided in~\Cref{app:implementation_detail}.
\begin{figure}[t]
    \centering
    \resizebox{0.8\columnwidth}{!}{
    \includegraphics[width=\textwidth,trim={6.5cm 0cm 6.5cm 7cm}, clip, page=7]{figures/figures_2.pdf}}
    \caption{Qualitative comparison of \ourmethod{} and previous top-down multimodal approaches on \datasetsecond{}. \ourmethod{} balances between coherence and quality of generated modality pairs.}
    \label{fig:results_celeb_comparison}
    \vspace{-10pt}
\end{figure}

\subsection{Multimodal Generation}
\label{sec:multimodal_exp}
\paragraph{Identity Bridge}
\label{sec:identity_bridge_exp}
Following prior works, we first evaluate the identity Bridge on \datasetfirst{}. This benchmark provides a simple multimodal case suitable for evaluating the identity bridge with a direct score switch. \Cref{fig:results_PolyMNIST} compares \ourmethod{} against prior approaches in both conditional and unconditional generation. \ourmethod{} prevailed in unconditional generation, striking a good balance between coherence and quality. In the conditional setting, it remains competitive with prior methods while attaining the highest image quality (lowest FID) with minimal loss in coherence. These results suggest that even a minimal bridge design is sufficient to capture shared structure in simple multimodal cases. Qualitative results further support these findings and are provided in \Cref{app:qualitative}.

\begin{wrapfigure}{r}{0.4\textwidth}
    \vspace{-20pt}
    \centering
    \includegraphics[width=\linewidth,trim={20cm 52cm 78cm 37cm},clip,page=5]{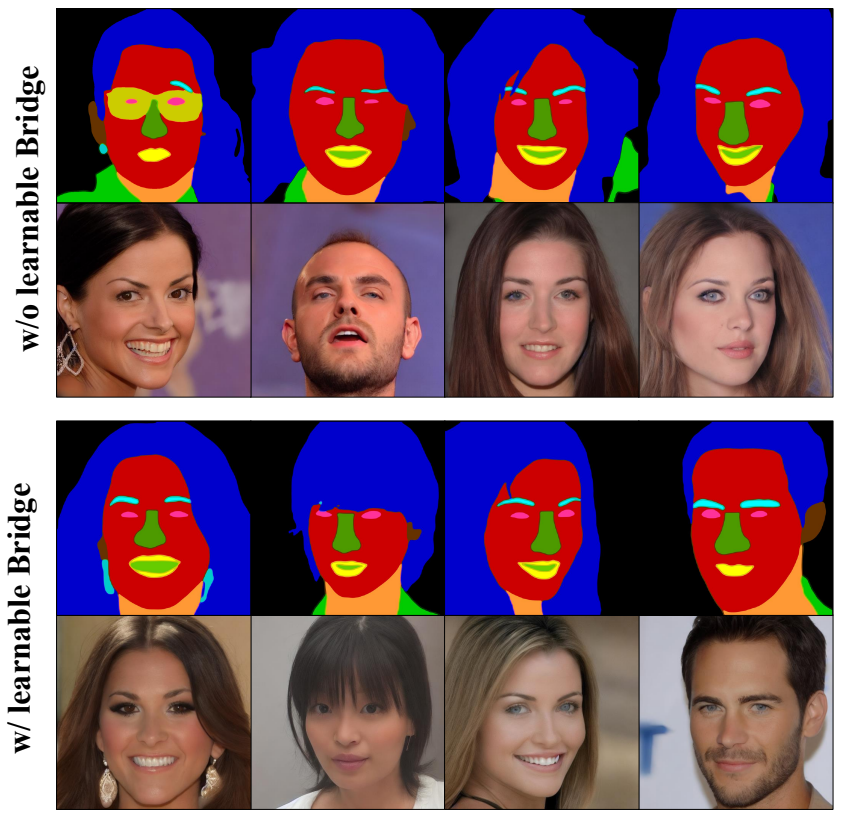}
    \caption{Images generated with a learnable bridge show superior results in terms of coherence over identity bridge in complex datasets.}
    \label{fig:results4}
\end{wrapfigure}

\begin{wraptable}{r}{0.6\textwidth}
    %\begin{table}
\centering
\caption{Results of multimodal generation on \datasetsecond{}. \ourmethod{} achieved the highest coherence without losing generation quality in conditional and unconditional generation. $\dagger$ and * refer to the used backbones: SD and ours, respectively.}
\label{tab:Celeb_main}
\resizebox{\linewidth}{!}{%
\begin{tabular}{l cc cc cc}
\toprule
\multirow{2}{*}{\textbf{Method}} & \multicolumn{2}{c}{\textbf{Joint}} & \multicolumn{2}{c}{\textbf{M$\to$F}} & \multicolumn{2}{c}{\textbf{F$\to$M}} \\
\cmidrule(lr){2-3} \cmidrule(lr){4-5} \cmidrule(l){6-7}
& \textbf{Coherence $\uparrow$} & \textbf{FID $\downarrow$} & \textbf{Coherence $\uparrow$} & \textbf{FID $\downarrow$} & \textbf{Coherence $\uparrow$} & \textbf{F1 $\uparrow$} \\
\midrule
MoPoE~\cite{elbo} & 0.59 & 151.23 & 0.50 & 163.83 & 0.51 & 0.72 \\
CMVAE~\cite{vae_diffusion} & 0.52 & 303.50 & 0.48 & 233.04 & 0.47 & 0.56 \\
MMVAE~\cite{mmvae} & 0.58 & 358.21 & 0.46 & 375.79 & 0.47 & 0.58 \\
MMVAE+~\cite{mmvae_plus} & 0.56 & 345.29 & 0.48 & 202.86 & 0.47 & 0.57 \\
MMVM VAE~\cite{mmvmvae} & 0.51 & 178.87 & \textbf{0.65} & 218.69 & 0.67 & 0.73 \\
MM-Diffusion$\dagger$~\cite{mmdiffusion} & 0.49 & 91.08 & 0.57 & 353.10 & 0.56 & 0.15 \\
MM-Diffusion* & 0.50 & 137.75 & 0.51 & 230.04 & 0.56 & 0.15 \\
MLD~\cite{mld} & 0.68 & 121.69 & \underline{0.64} & 227.00 & 0.71 & 0.76 \\
\midrule
\ourmethod{} ($t_{split} = 10\%$) & 0.49 & 60.24 & 0.49 & 200.84 & 0.49 & 0.54 \\
\ourmethod{} ($t_{split} = 20\%$) & 0.52 & \textbf{59.22} & 0.49 & 201.62 & 0.54 & 0.62 \\
\ourmethod{} ($t_{split} = 30\%$) & 0.60 & 59.73 & 0.50 & \textbf{65.50} & 0.62 & 0.76 \\
\ourmethod{} ($t_{split} = 40\%$) & 0.67 & 60.30 & 0.55 & \underline{66.83} & 0.68 & 0.82 \\
\ourmethod{} ($t_{split} = 50\%$) & 0.73 & 60.84 & 0.60 & 107.22 & 0.72 & 0.85 \\
\ourmethod{} ($t_{split} = 60\%$) & 0.76 & 60.20 & 0.59 & 155.90 & 0.75 & \underline{0.87} \\
\ourmethod{} ($t_{split} = 70\%$) & \underline{0.78} & 59.47 & 0.55 & 208.05 & \underline{0.76} & \textbf{0.88} \\
\ourmethod{} ($t_{split} = 80\%$) & \textbf{0.79} & \underline{59.23} & 0.55 & 180.58 & \textbf{0.77} & 0.82 \\
\ourmethod{} ($t_{split} = 90\%$) & \underline{0.78} & 59.41 & 0.56 & 195.57 & \underline{0.76} & 0.81 \\
\bottomrule
\end{tabular}}
    \vspace{-5pt}
\end{wraptable}
\vspace{-12pt}
\paragraph{Learnable Bridge}
\label{sec:learnable_bridge_exp}
We evaluate \ourmethod{} with a learnable bridge on \datasetsecond{}. \textit{Note that} here, $t_{split}$ denotes the percentage of the trajectory shared before the split. For instance, $t_{split}=30\%$ means that the first $30\%$ of the diffusion process is shared across modalities. As shown in \Cref{tab:Celeb_main}, \ourmethod{} achieves the best image quality (FID) across all $t_{split}$ values while maintaining strong coherence, consistently outperforming both VAE- and diffusion-based baselines. In contrast, VAE methods achieve higher coherence at the cost of image quality, while MM-Diffusion improves fidelity but struggles to maintain alignment. In conditional generation (face$\rightarrow$mask and mask$\rightarrow$face), \ourmethod{} achieves comparable performance, achieving high-quality synthesis with accurate structural alignment, as reflected by competitive F1 scores for mask prediction. Qualitatively (\Cref{fig:results_celeb_comparison,fig:celeb_conditional}), \ourmethod{} produces realistic samples with clear correspondence between faces and masks, significantly outperforming top-down baselines. Additional examples are provided in \Cref{app:qualitative}.
\begin{wrapfigure}{r}{0.4\textwidth} 
    \centering
    \vspace{-10pt}
    \includegraphics[width=\linewidth,trim={2.8cm 9cm 22cm 5cm},clip,page=4]{figures/figures_2.pdf}
    \caption{conditional generation of \ourmethod{} on \datasetsecond{}.}
    \label{fig:celeb_conditional}
    \vspace{-10pt}
\end{wrapfigure}
\paragraph{Ablation Studies}
\label{sec:ablation}
We ablate two main components in the learnable bridge: the bridge network $\mathcal{B}_\theta$ and the noise injector. The results are shown in~\Cref{tab:Celeb_ablation}. While the FID score for unconditional generation remains stable (around 60) across all ablations, the coherence changes significantly. Removing the bridge network (identity bridge) causes a substantial decrease in coherence, indicating that without it, the generated samples are not well-correlated even if they are realistic; this finding is illustrated in~\Cref{fig:results4}. These results clearly show the need for learnable bridges in complex data structures compared to a simple change of score function at $t_{split}$. The noise injector primarily impacts conditional generation, as the FID increases when it is omitted in the mask$\rightarrow$face case. We further ablate the capacity of $\mathcal{B}_\theta$ across small, medium, and large variants in~\Cref{tab:bridge_capacity_ablation} by increasing and depth and base channels of the U-net. Increasing the size of the bridge consistently improves coherence while maintaining strong FID, suggesting that a larger capacity places the trajectory in a better initialization for the second model. These findings show that bridge design choices significantly affect performance.

\begin{wrapfigure}{r}{0.4\textwidth} % 'r' for right side, 50% of text width
    \centering
    \includegraphics[width=\linewidth,trim={0cm 0cm 0cm 0cm},clip,page=2]{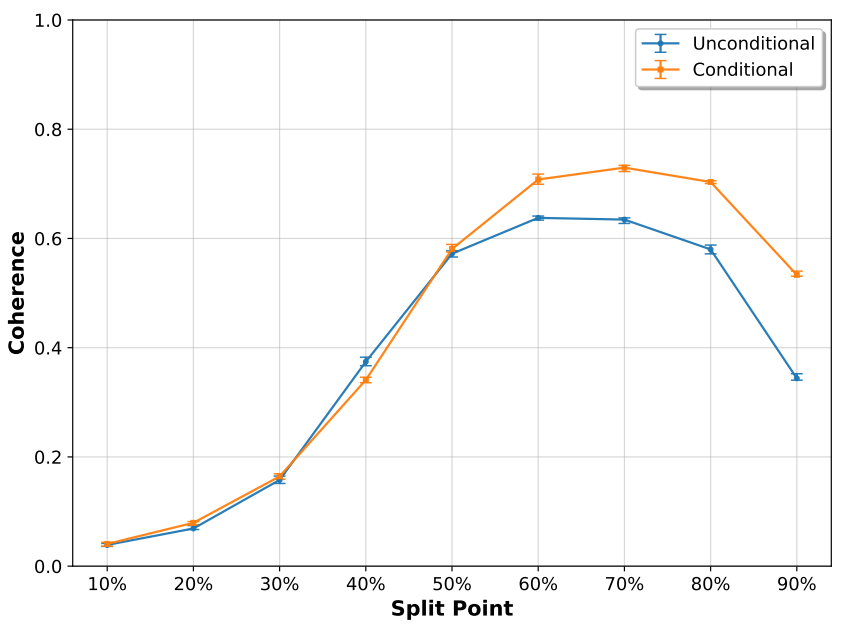}
    \caption{Sensitivity analysis on $t_{split}$ on \datasetfirst{}}
    \label{fig:PolyMNIST_sensitivity}
\end{wrapfigure}

\begin{wraptable}{r}{0.5\textwidth}
    \vspace{-10pt}
    %%%%%%%%%%%%%%%%%%%%%%%%%%%%%%%%%%%%%%%%%%%%%
%\begin{table}[h]
\centering
\caption{Ablation study of \ourmethod{}. Br. and N.I. denote the bridge and noise injector. Coh. is coherence.}
\label{tab:Celeb_ablation}
\resizebox{\linewidth}{!}{%
\begin{tabular}{lc c c c c c c}
\toprule
\multicolumn{2}{c}{\textbf{Ablation}} & \multicolumn{2}{c}{\textbf{Joint}} & \multicolumn{2}{c}{\textbf{M$\to$F}} & \multicolumn{2}{c}{\textbf{F$\to$M}} \\
\cmidrule(r){1-2} \cmidrule(lr){3-4} \cmidrule(lr){5-6} \cmidrule(l){7-8}
\textbf{Br.} & \textbf{N.I.} & \textbf{Coh. $\uparrow$} & \textbf{FID $\downarrow$} & \textbf{Coh. $\uparrow$} & \textbf{FID $\downarrow$} & \textbf{Coh. $\uparrow$} & \textbf{F1 $\uparrow$} \\
\midrule
$\checkmark$ & $\checkmark$ & 0.73 & 60.84 & \textbf{0.60} & 107.22 & \textbf{0.72} & \textbf{0.85} \\
$\checkmark$ & $\times$ & \textbf{0.76} & \textbf{59.89} & 0.50 & 216.39 & 0.70 & 0.78 \\
$\times$ & $\checkmark$ & 0.48 & 60.82 & 0.50 & \textbf{52.03} & 0.48 & 0.53 \\
$\times$ & $\times$ & 0.48 & 59.90 & 0.50 & 292.58 & 0.49 & 0.02 \\
\bottomrule
\end{tabular}}
%\end{table}

    \vspace{-5pt}
\end{wraptable}
\paragraph{Sensitivity Analysis on $t_{split}$}
A natural question that arises in \ourmethod{} is: \textit{when should the diffusion trajectories be split, and how does the choice of $t_{split}$ affect generation quality and coherence?} We answer these questions by conducting a sensitivity analysis by varying the split point $t_{split}$ across the reverse process on both \datasetfirst{} and \datasetsecond{}. The corresponding results are shown in~\Cref{tab:Celeb_main} and ~\Cref{fig:PolyMNIST_sensitivity}. The coherence shows a clear trend: \textit{later splits (higher $t_{split}$) lead to higher coherence, meaning that the generated representations are more correlated with each other}. This behavior aligns with our core hypothesis regarding shared latent representations. A larger shared segment of the reverse trajectory (e.g., $t_{split}=70\%$) allows both modalities to co-evolve for a longer duration, enhancing their structural alignment. Conversely, earlier splits (e.g., $t_{split}=30\%$) reduce the overlap in shared latent evolution, leading to weaker correspondence. Moreover, the coherence drops after a certain point (around $t_{split}=70\%$) given that more steps are needed to move the trajectory to the new modality distribution.

\subsection{Scalability and Flexibility of Bottom-up \ourmethod{}}
\label{sec:scalability}
\begin{figure}
    \centering
    \includegraphics[width=\linewidth,trim={5.5cm 7.5cm 6cm 7cm},clip,page=1]{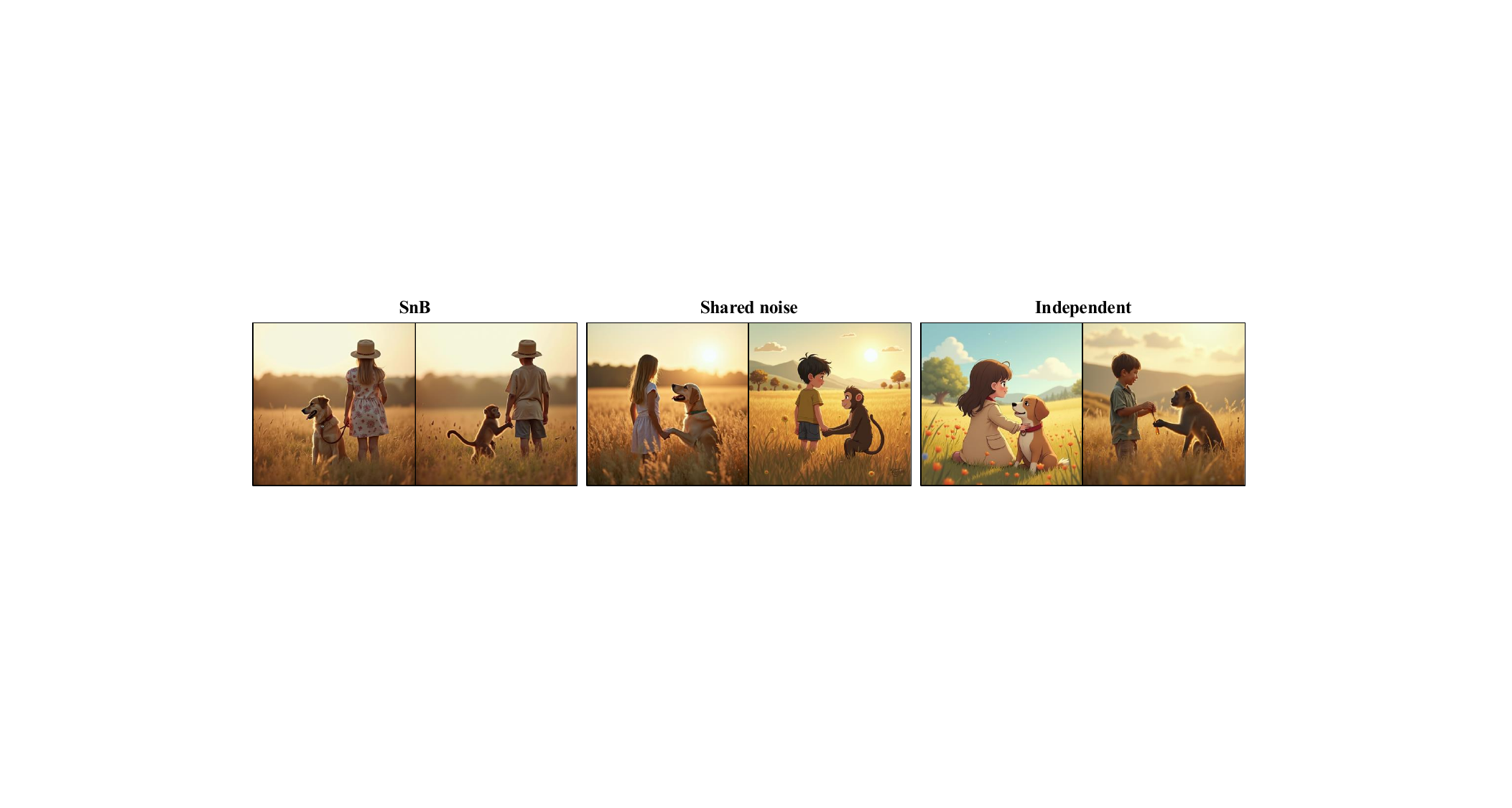}
    \caption{Interleaved co-generation on \datasetthird{}. \ourmethod{} shows the most correlated co-generated pairs. Prompt 1: a \textcolor{red}{girl} and her \textcolor{red}{dog} in a field. Prompt 2: a \textcolor{blue}{boy} and his \textcolor{blue}{monkey} in a field.}
    \vspace{-10pt}
    \label{fig:interleaved_cogen}
\end{figure}

We analyze \ourmethod{}'s scalability through leveraging large foundation diffusion models~\cite{flux,sdxl,ldm} on downstream tasks. We evaluate on \datasetthird{} using the identity bridge with a prompt swap at $t_{split}$ as the score intervention. We consider two tasks, \textit{interleaved co-generation}, where paired images are generated from two prompts with shared structure, and \textit{image editing}, where an input image is noised to $t_{split}$ and then denoised under a new prompt. We compare against two baselines, independent diffusion models per prompt (no shared noise) and shared-noise diffusion models (could be considered \ourmethod{} with $t_{split}=0\%$). \Cref{tab:Pie_bench_main} shows that across multiple backbones, all methods achieve comparable FID and CLIP~\cite{clip} scores, indicating similar image quality and prompt alignment. However, \ourmethod{} achieves the highest pair consistency, producing semantically aligned pairs while preserving prompt-specific variations. This advantage is clearly visualized in \Cref{fig:interleaved_cogen,fig:image_editing}, where the generated images are aligned and representing the prompts. These results highlight that \ourmethod{} enables coherent multi-output generation at scale, acting as a plug-and-play framework for foundation models. Additional details and visualizations are provided in \Cref{app:downstream}.

\begin{figure}[H]
    \centering
    \includegraphics[width=\linewidth,trim={0.5cm 11.2cm 15.5cm 3.2cm},clip,page=2]{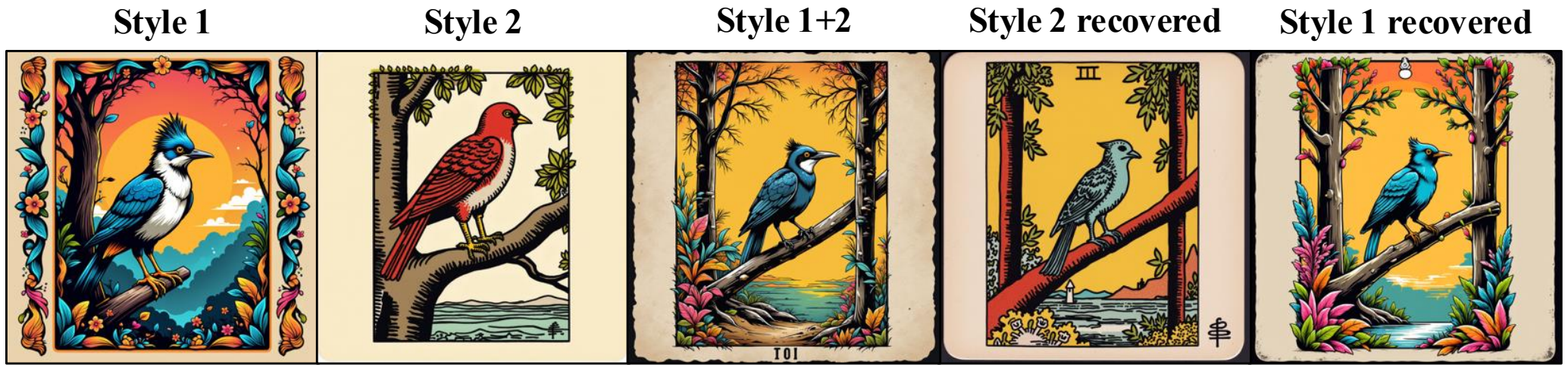}
    \caption{Qualitative \datasetthird{} style results. \ourmethod{} demonstrates compositional abilities in combining and separating modalities. Prompt: a bird standing on tree branch.}
    \label{fig:style}
\end{figure}

\begin{wraptable}{r}{0.48\textwidth}
    \vspace{-10pt}
    %\begin{table}[h]
\centering
\caption{Ablation study on the learnable bridge ($\mathcal{B}_\theta$) capacity on \datasetsecond{} at $t_{\text{split}} = 50\%$. Medium denotes the default configuration used in all other experiments.}
\label{tab:bridge_capacity_ablation}
\setlength{\tabcolsep}{3.5pt}
\resizebox{\linewidth}{!}{%
\small
\begin{tabular}{@{}lcccc|cc@{}}
\toprule
\multicolumn{1}{@{}c}{\textbf{Bridge}} & & \textbf{Base} & \textbf{Params} & \textbf{Mem.} & & \\
\multicolumn{1}{@{}c}{\textbf{Variant}} & \multirow{-2}{*}{\textbf{Depth}} & \textbf{Channels} & \textbf{(M)} & \textbf{(MB)} & \multirow{-2}{*}{\textbf{Coh.} $\uparrow$} & \multirow{-2}{*}{\textbf{FID} $\downarrow$} \\
\midrule
Small  & 2 & 16 & 2.1   & 8.37   & 0.68 & \textbf{60.14} \\
Medium & 4 & 32 & 15.8  & 60.29  & 0.73 & 60.84 \\
Large  & 6 & 64 & 233.3 & 890.18 & \textbf{0.76} & 60.72 \\
\bottomrule
\end{tabular}
}
%\end{table}
    \vspace{-10pt}
\end{wraptable}
Another form of score intervention is through parameter modification (\Cref{sec:methods}). Leveraging the availability of task-specific fine-tuned variants of diffusion models (e.g., LoRA~\cite{lora}), we demonstrate the compositionality of \ourmethod{} on the style transfer subset of \datasetthird{}. We apply two distinct LoRA styles\footnote{https://huggingface.co/multimodalart/flux-tarot-v1}\footnote{https://huggingface.co/SouthbayJay/t-shirt-vector-style-flux} at different timesteps, introducing two intervention points. Starting from the base model, we activate style 1 at $t_{style}^{(1)}$ and subsequently style 2 at $t_{style}^{(2)}$. This type of score modification results in blended images that exhibit both styles. Remarkably, we demonstrate that this process is reversible, where individual styles can be decoupled from their combination, as shown in \Cref{fig:style}. These results highlight the ability of \ourmethod{} to compose and control multiple model adaptations within a unified sampling process. Additional results and details are provided in \Cref{app:downstream}.
\begin{table}[t]
\centering
%\vspace{-10pt}
\caption{Quantitative evaluation on PIE-Bench for interleaved co-generation and image editing tasks.}
\label{tab:Pie_bench_main}
\resizebox{\textwidth}{!}{%
\begin{tabular}{ll cccccccc ccccc}
\toprule
\multirow{3}{*}{\rotatebox{90}{\textbf{Model}}} & \multirow{3}{*}{\textbf{\ourmethod{} Variant}} & \multicolumn{8}{c}{\textbf{Interleaved Co-Generation}} & \multicolumn{5}{c}{\textbf{Image Editing}} \\
\cmidrule(lr){3-10} \cmidrule(l){11-15}
& & \multicolumn{2}{c}{\textbf{Target}} & \multicolumn{2}{c}{\textbf{Source}} & \multicolumn{4}{c}{\textbf{Consistency}} & \multicolumn{5}{c}{\textbf{Target}} \\
\cmidrule(lr){3-4} \cmidrule(lr){5-6} \cmidrule(lr){7-10} \cmidrule(l){11-15}
& & FID $\downarrow$ & CLIP $\uparrow$ & FID $\downarrow$ & CLIP $\uparrow$ & FID $\downarrow$ & LPIPS $\downarrow$ & MSE $\downarrow$ & PSNR $\uparrow$ & FID $\downarrow$ & LPIPS $\downarrow$ & MSE $\downarrow$ & PSNR $\uparrow$ & CLIP $\uparrow$ \\
\midrule
\multirow{3}{*}{\rotatebox{90}{\begin{tabular}{c} Flux \\ \cite{flux} \end{tabular}}}
& Independent & 76.60 & 32.58 & 75.27 & 32.18 & 61.93 & 0.67 & 0.10 & 10.00 & - & - & - & - & - \\
& shared noise & 76.09 & 32.38 & 74.74 & 32.18 & 56.32 & 0.48 & 0.03 & 14.76 & 78.24 & 0.76 & 0.15 & 8.25 & 32.52 \\
& \ourmethod{} & \textbf{75.95} & 31.79 & 74.86 & 32.23 & \textbf{51.69} & \textbf{0.39} & \textbf{0.02} & \textbf{17.03} & 76.94 & 0.70 & 0.09 & 10.55 & 32.45 \\
\midrule
\multirow{3}{*}{\rotatebox{90}{\begin{tabular}{c} SDXL \\ \cite{sdxl} \end{tabular}}}
& Independent & 77.61 & \textbf{33.59} & \textbf{74.14} & \textbf{32.63} & 70.96 & 0.69 & 0.08 & 11.03 & - & - & - & - & - \\
& shared noise & 79.03 & 33.39 & 74.59 & 32.57 & 68.16 & 0.50 & 0.03 & 15.02 & 110.41 & 0.71 & 0.11 & 9.77 & 31.30 \\
& \ourmethod{} & 77.49 & 33.32 & 74.49 & 32.54 & 65.53 & 0.45 & \textbf{0.02} & 16.59 & 114.47 & 0.69 & 0.09 & 10.46 & 30.96 \\
\midrule
\multirow{3}{*}{\rotatebox{90}{\begin{tabular}{c} SD 2.1 \\ \cite{ldm} \end{tabular}}}
& Independent & 81.82 & 33.27 & 75.61 & 32.54 & 75.36 & 0.77 & 0.10 & 9.94 & - & - & - & - & - \\
& shared noise & 82.07 & 33.13 & 76.13 & 32.39 & 71.47 & 0.54 & 0.04 & 13.99 & 86.94 & 0.69 & 0.09 & 10.65 & 31.69 \\
& \ourmethod{} & 79.93 & 32.70 & 76.89 & 32.45 & 67.00 & 0.48 & 0.03 & 15.10 & 88.57 & 0.66 & 0.07 & 11.44 & 31.53 \\
\midrule
\multirow{3}{*}{\rotatebox{90}{\begin{tabular}{c} SD 1.5 \\ \cite{ldm} \end{tabular}}}
& Independent & 79.88 & 33.05 & 76.69 & 32.42 & 72.84 & 0.78 & 0.13 & 8.85 & - & - & - & - & - \\
& shared noise & 80.22 & 32.91 & 75.54 & 32.28 & 68.98 & 0.55 & 0.05 & 12.88 & 77.19 & 0.67 & 0.08 & 11.07 & \textbf{32.73} \\
& \ourmethod{} & 78.77 & 32.66 & 77.00 & 32.46 & 67.28 & 0.49 & 0.04 & 13.93 & \textbf{75.54} & \textbf{0.63} & \textbf{0.06} & \textbf{11.99} & 32.50 \\
\bottomrule
%\vspace{-20pt}
\end{tabular}%
}
\end{table}

\paragraph{Downstream Application: Zero-Shot Multimodal MRI Synthesis}
Medical imaging data is inherently multimodal, for example, MRI scans provide complementary diagnostic views across multiple contrast levels. Multimodal MRI synthesis is a challenging problem due to the difficulty of collecting paired multi-contrast scans at scale. Existing top-down multimodal generation methods typically require jointly aligned training data across all modalities, which is expensive and often impractical in real-world medical settings. At the same time, recent medical foundation diffusion models have demonstrated strong capabilities in generating high-quality MRI images from text prompts, motivating the question of \textit{whether such pretrained unimodal generators can be composed into a multimodal generation framework without retraining}.

To study this setting, we apply \ourmethod{} in a zero-shot manner for multimodal MRI synthesis using \datasetfourth{} and the pretrained medical diffusion model Prompt2MedImage\footnote{https://huggingface.co/Nihirc/Prompt2MedImage}. We consider four common MRI contrasts: T1, T2, FLAIR, and DWI, each capturing complementary physiological properties of the same anatomy. Using the identity bridge with prompt-based score intervention, \ourmethod{} generates different MRI contrasts from a partially shared diffusion trajectory while preserving the underlying anatomical structure.

\begin{wrapfigure}{r}{0.5\textwidth} 
    \centering
    \vspace{-12pt}
    \includegraphics[width=\linewidth,trim={0.5cm 7.5cm 23cm 7cm},clip,page=5]{figures/figures_2.pdf}
    \caption{Editing on \datasetthird{}. a \textcolor{red}{gray horse} in the field$\rightarrow$ a \textcolor{blue}{black cow with white patterns} in the field.}
    \vspace{-10pt}
    \label{fig:image_editing}
\end{wrapfigure}
\begin{wrapfigure}{r}{0.55\textwidth}
    \centering
    \vspace{-20pt}
    \includegraphics[width=\linewidth,trim={1cm 2cm 10.5cm 1cm},clip,page=3]{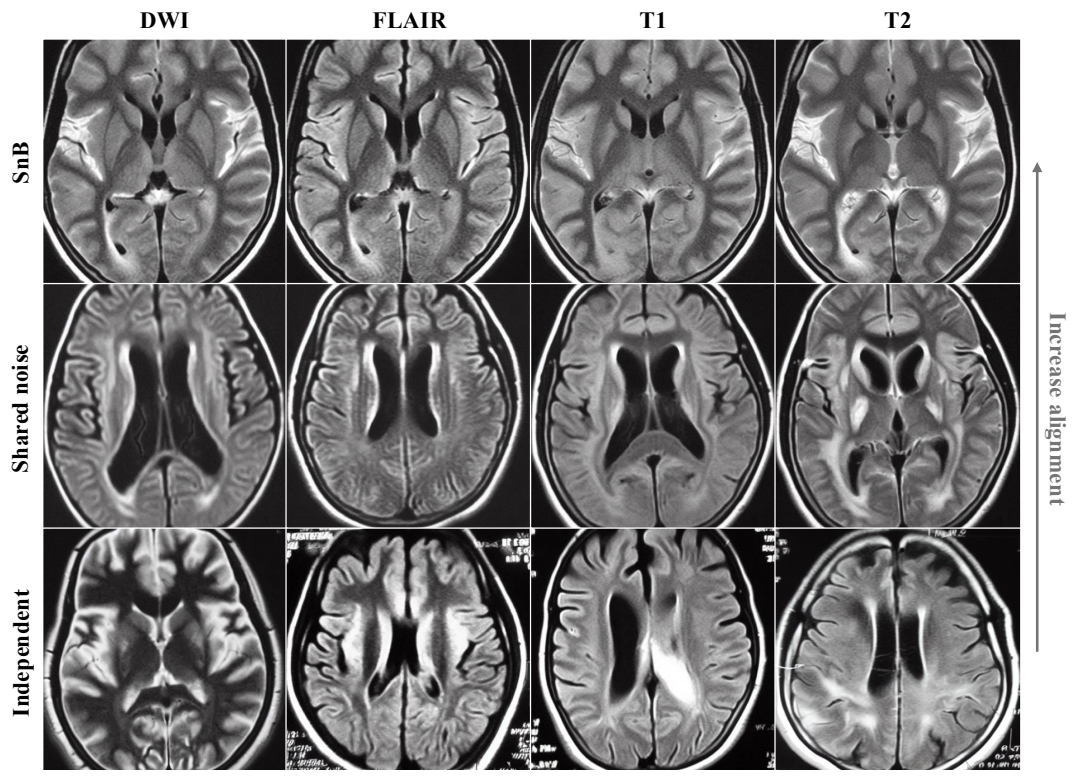}
    \caption{Independent models lead to completely different structures, sharing the noise component increases alignment, while \ourmethod{} is the most aligned.}
    \label{fig:medical_main}
    \vspace{-30pt}
\end{wrapfigure}

We evaluate on a test subset spanning brain, joint, and spine MRI scans. This setting directly tests whether \ourmethod{} can coordinate independently generated modalities without paired multimodal training. As shown in \Cref{tab:ROCO_main}, all methods achieve similar FID and PubMedCLIP~\cite{medclip} scores due to the shared pretrained backbone. However, \ourmethod{} consistently achieves stronger modality consistency, indicating improved structural alignment across generated contrasts. Qualitative examples in \Cref{fig:medical_main} further demonstrate that \ourmethod{} preserves consistent anatomy while producing clear contrast-specific appearance changes across modalities. For example, the prompt `T2 axial brain magnetic resonance imaging scan showing hypointense rim with hyperintense heterogenous core in the left frontoparietal region'' can be transformed into different modalities by replacing `T2'' with `T1'', `FLAIR'', or ``DWI''.
\begin{table}[t]
\centering
\caption{Quantitative results across three medical imaging categories of \datasetfourth{}. Modalities not present in a specific anatomy are marked with a dash (-)}
\label{tab:ROCO_main}
\resizebox{\textwidth}{!}{%
\begin{tabular}{ll ccccc ccccc ccc}
\toprule
\multirow{2}{*}{\textbf{Anatomy}} & \multirow{2}{*}{\textbf{Split Point}} & \multicolumn{5}{c}{\textbf{PubMedCLIP~\cite{medclip} $\uparrow$}} & \multicolumn{5}{c}{\textbf{FID $\downarrow$}} & \multicolumn{3}{c}{\textbf{Consistency}} \\
\cmidrule(lr){3-7} \cmidrule(lr){8-12} \cmidrule(l){13-15}
& & T1 & T2 & FLAIR & DWI & PD & T1 & T2 & FLAIR & DWI & PD & MSE $\downarrow$ & PSNR $\uparrow$ & LPIPS $\downarrow$ \\
\midrule
\multirow{3}{*}{Brain} 
& Independent & 28.87 & \textbf{29.42} & 30.58 & \textbf{27.43} & - & 137.37 & 152.98 & 145.25 & 134.90 & - & 0.13 & 8.79 & 0.60 \\
& Shared Noise & \textbf{28.95} & 29.28 & \textbf{30.70} & 27.38 & - & 137.57 & 154.88 & 138.01 & 132.59 & - & 0.04 & 14.06 & 0.38 \\
& \ourmethod{} & 28.85 & 29.17 & 30.24 & 27.08 & - & \textbf{134.67} & \textbf{147.44} & \textbf{134.37} & \textbf{132.46} & - & \textbf{0.02} & \textbf{17.90} & \textbf{0.27} \\
\midrule
\multirow{3}{*}{Hip/Shld/Elbow} 
& Independent & 30.96 & 31.15 & - & - & 30.13 & 258.82 & 256.06 & - & - & 268.05 & 0.16 & 7.93 & 0.61 \\
& Shared Noise & \textbf{31.38} & \textbf{31.34} & - & - & 30.04 & \textbf{247.53} & \textbf{255.18} & - & - & \textbf{254.40} & 0.04 & 14.60 & 0.34 \\
& \ourmethod{} & 30.81 & 31.10 & - & - & \textbf{30.19} & 258.76 & 262.59 & - & - & 258.97 & \textbf{0.02} & \textbf{18.09} & \textbf{0.24} \\
\midrule
\multirow{3}{*}{Spine} 
& Independent & 31.18 & \textbf{31.17} & - & - & - & 264.25 & 269.55 & - & - & - & 0.14 & 8.64 & 0.62 \\
& Shared Noise & 31.09 & 30.88 & - & - & - & \textbf{263.34} & \textbf{263.08} & - & - & - & 0.02 & 17.77 & 0.24 \\
& \ourmethod{} & \textbf{31.25} & 31.02 & - & - & - & 267.65 & 270.08 & - & - & - & \textbf{0.01} & \textbf{21.36} & \textbf{0.16} \\
\bottomrule
%\vspace{-30pt}
\end{tabular}%
}
\end{table}
\vspace{-5pt}
\section{Conclusion}
\vspace{-5pt}
\label{sec:conclusion}
%\begin{wrapfigure}{r}{0.7\textwidth}

We introduced \ourmethod{}, a bottom-up framework for multimodal generation that composes pretrained unimodal diffusion models during sampling. Unlike prior top-down approaches that rely on learning a shared latent representation through joint training, \ourmethod{} couples modality-specific trajectories. This design scales with foundation diffusion models and enables direct reuse of strong pretrained experts.

\ourmethod{} empirically achieves a favorable balance between coherence and sample quality. Beyond standard multimodal generation, we demonstrate its flexibility through zero-shot extensions to downstream tasks by leveraging foundation models, highlighting a level of scalability and practicality not accessible to traditional approaches. Our results in medical imaging further illustrate the potential of this paradigm in real-world settings, suggesting that effective multimodal generation benefits from composable and scalable designs.

%\begin{wrapfigure}{r}{0.7\textwidth}

\paragraph{Limitations}
Despite its flexibility, \ourmethod{} introduces new design choices. In particular, selecting an appropriate bridge mechanism and $t_{split}$ may require empirical tuning depending on the modality gap and data complexity. Additionally, performance can be sensitive to the choice of the initial trajectory, as some modalities provide more suitable shared representations than others (see \Cref{app:start_point}). Finally, \ourmethod{} assumes access to pretrained unimodal models for each modality, which may be costly in some settings. However, this requirement is mitigated in practice by the growing availability of large pretrained and adaptable foundation models (see \Cref{sec:scalability}).

\paragraph{Future Work}
Future directions include extending \ourmethod{} beyond visual modalities, handling larger semantic gaps between modalities, and developing robust strategies for selecting split points and bridge designs. More broadly, we view the bridge as a modular component, and further exploration of its design and its interaction with diffusion trajectories is a promising avenue. We hope this work encourages a shift toward bottom-up composable approaches for scalable multimodal generation.
\section{acknowledgments}
\label{sec:acknowledgments}
This material is based upon work supported by the National Science Foundation (NSF) under Grant Nos. 2218063 and 2551626. Computations were supported through the AMD University Program AI \& HPC Cluster.

\bibliographystyle{abbrv}
\bibliography{ref.bib}

\newpage
\appendix
% Source - https://tex.stackexchange.com/a/121055
% Posted by Andrew Swann, modified by community. See post 'Timeline' for change history
% Retrieved 2026-05-04, License - CC BY-SA 4.0
\crefalias{section}{appendix}
\section{Noise Injector in $\mathcal{B}_\mathbf{\theta}$}
\label{app:noise_injector}
A Mean Squared Error (MSE) objective essentially cannot model stochastic variations, as MSE minimization leads to the conditional expectation $\mathbb{E}[\mathcal{B}_\mathbf{\theta}(\mathbf{x}_{t_{split}}^{(1)}) \ |\ \hat{\mathbf{x}}_{t_{split}}^{(1)}]$ as the optimal predictor. Given that $\mathbb{E}[\epsilon_{t_{split}}]=\textbf{0}$, the model would never learn noisy latents across the reverse process, resulting in smooth deterministic outputs. To map $\mathbf{x}_{t_{split}}^{(1)}$ to the correct corresponding state $\mathbf{x}_{t_{split}}^{(2)}$ after bridging, we introduce a \textbf{noise injector} that adds noise back into $\hat{\mathbf{x}}_t^{(2)}$. The amount of noise added is determined by the diffusion coefficient $g_2(t)$ of the forward process in~\Cref{eq:SDE_forward}. Practically, this is controlled by the scheduler of the diffusion model. In this section, we provide a formal derivation showing why an MSE loss fails to learn the stochastic mapping between noisy latents, causing the model to converge to the conditional mean predictor and produce non-noisy results.

Let $\mathbf{x}_t^{(1)}$ and $\mathbf{x}_t^{(2)}$ be the noisy latents of their respective modalities at timestep $t$. For notation simplicity, we consider the noise added by the scheduler at $t$ as $\epsilon_t$, where $\mathbf{x}_t^{(2)} = \gamma_t \mathbf{x}_0^{(2)} + \epsilon_t$, $\gamma_t \text{ and } \epsilon_t$ are determined by the scheduler. The standard MSE objective is given as follows,
\begin{align}
\begin{split}
\label{eq:mse}
     \mathcal{L}_{\text{MSE}} &= \mathbb{E}_{t, \mathbf{x}_t^{(1)}, \mathbf{x}_t^{(2)}} \| \mathcal{B}_\mathbf{\theta}(\mathbf{x}_t^{(1)}, t) - \mathbf{x}_t^{(2)} \|_2^2 \\
    &= \mathbb{E}_{t, \mathbf{x}_t^{(1)}} \left[ \mathbb{E}_{\mathbf{x}_0^{(2)}} \left[ \| \mathcal{B}_\mathbf{\theta}(\mathbf{x}_t^{(1)}, t) - (\gamma_t\mathbf{x}_0^{(2)} + \epsilon_t) \|_2^2 \right] \right]
\end{split}
\end{align}
For any fixed input $(\mathbf{x}_t^{(1)}, t)$, the model predicts a deterministic output $\hat{\mathbf{x}_t}^{(2)} =\mathcal{B}_\mathbf{\theta}(\mathbf{x}_t^{(1)}, t)$, we can extend the inner expectation as follows, 
\begin{equation}
\begin{gathered}
 \mathbb{E}_{\mathbf{x}_0^{(2)}} \left[ \| \hat{\mathbf{x}}_t^{(2)} - (\gamma_t\mathbf{x}_0^{(2)} + \epsilon_t) \|_2^2 \right] \\
= \mathbb{E}_{\mathbf{x}_0^{(2)}} \left[ \| (\hat{\mathbf{x}}_t^{(2)} - \gamma_t\mathbf{x}_0^{(2)}) - \epsilon_t \|_2^2 \right] \\
= \mathbb{E}_{\mathbf{x}_0^{(2)}} \left[ \| \hat{\mathbf{x}}_t^{(2)} - \gamma_tx_0^{(2)} \|_2^2 - 2 (\hat{\mathbf{x}}_t^{(2)} - \gamma_t\mathbf{x}_0^{(2)})^\top \epsilon_t + \| \epsilon_t \|_2^2 \right] \\
 = \| \hat{\mathbf{x}}_t^{(2)} - \gamma_t\mathbf{x}_0^{(2)} \|_2^2 - 2 (\hat{\mathbf{x}}_t^{(2)} - \gamma_t\mathbf{x}_0^{(2)})^\top \mathbb{E}_{\mathbf{x}_0^{(2)}}[\epsilon_t] + \mathbb{E}_{\mathbf{x}_0^{(2)}}[\| \epsilon_t \|_2^2] \\
= \| \hat{\mathbf{x}}_t^{(2)} - \gamma_tx_0^{(2)} \|_2^2 + \sigma_t^2
\end{gathered}
\end{equation}
Where $\mathbb{E}[\epsilon_t] = 0$ and $\mathbb{E}[\epsilon_t^\top \epsilon_t] = \sigma_t^2$ ($\sigma^2_t$ is the variance of the added Gaussian distribution at $t$). Since $\sigma_t^2 \text{ and } \gamma_t$ are constants with respect to $\hat{\mathbf{x}}_t^{(2)}$, the loss is minimized when $\| \hat{\mathbf{x}}_t^{(2)} - \mathbf{x}_0^{(2)} \|_2^2$ is minimized. Therefore, for any input $\mathbf{x}_t^{(1)}$, the optimal prediction under the MSE loss is the clean latent $\hat{\mathbf{x}}_t{^*}^{(2)} = \mathbf{x}_0^{(2)}$. This proves that the MSE-optimal bridge network ignores all noise terms added and predicts the conditional expectation $\mathbb{E}[\mathbf{x}_t^{(2)} \mid \mathbf{x}_t^{(1)}] = \mathbf{x}_0^{(2)}$, failing to learn the true noisy components in the reverse process.

In practice, we treat the noise injection strength as a hyperparameter and select it empirically. It is typically set to correspond to reverting the reverse process by approximately $10\%$ to $30\%$ of the diffusion trajectory from $t_{split}$. We note that it had higher impact on conditional generation as can be shown in \Cref{tab:Celeb_ablation}.
%%%%%%%%%%%%%%%%%%%%%%%%%%%%%%%%%%%%%%%%%%%%%%%%%%%%%%%%%%%%%%%%%%%%%%%%%%%%%%
\section{Implementation Details}
\label{app:implementation_detail}
\subsection{Datasets} 
We report multimodal experimental results on multiple benchmarks. \datasetfirst{} and \datasetsecond{} are two well-established datasets for multimodal generation, allowing both unconditional joint and conditional generation. \datasetfirst{} is one of the first datasets proposed in this field, and it consists of five visual modalities of digits with varying backgrounds. Generally, this benchmark is low in complexity due to the lower resolution of images; however, it serves as a great way to measure a model's capability in joint generation across a relatively high number of modalities, five. 

\begin{figure}[h]
    \centering
    \resizebox{0.8\columnwidth}{!}{
    \includegraphics[width=\textwidth,trim={1cm 4.2cm 0.2cm 4cm},clip,page=2]{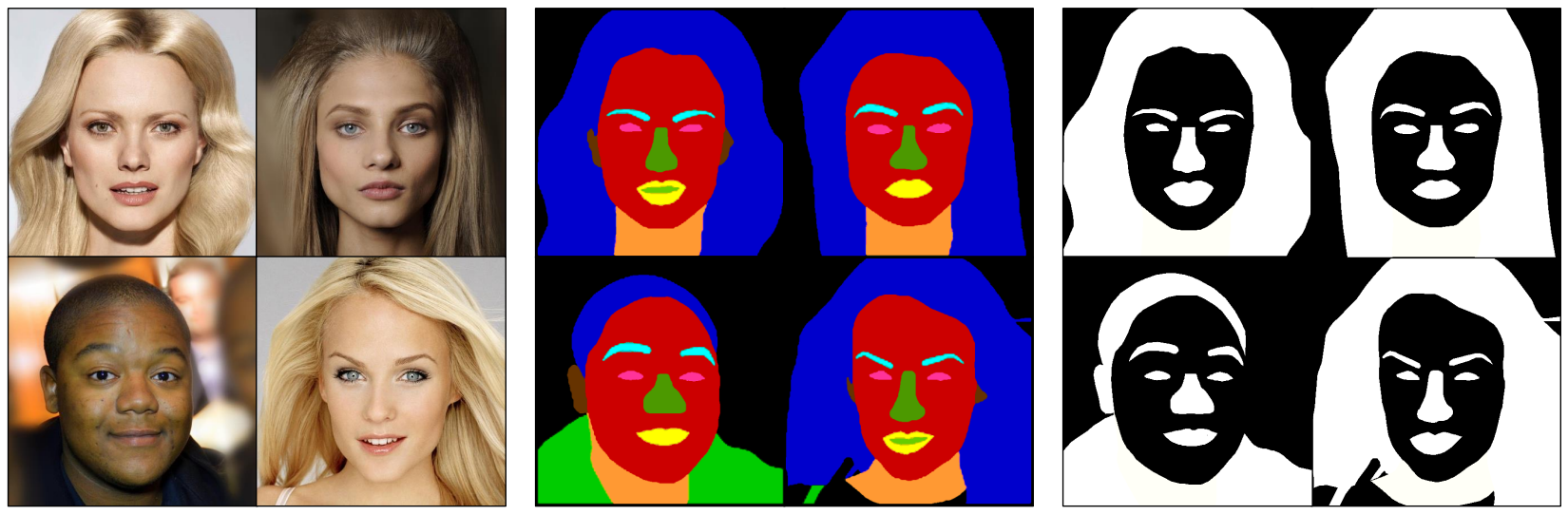}}
    \caption{A sample of \datasetfirst{}. Unlike prior works that use binary segmentation masks (right samples), we use a more comprehensive mask with 15 classes (center samples)~\cite{mld,mvae}.}
    \label{fig:celeb_sample}
\end{figure}

\datasetsecond{}, on the other hand, is much more complex and higher in resolution. It contains $30,000$ paired faces and segmentation masks at 512×512 resolution, we use 15-class masks, extending beyond the binary masks used in prior works~\cite{mld,mvae}. \Cref{fig:celeb_sample} shows a batch of samples from \datasetfirst{} alongside the binary-mask variant from~\cite{mld,mvae} for comparison.

\begin{figure}[h]
    \centering
    \resizebox{0.7\columnwidth}{!}{
    \includegraphics[width=\textwidth,trim={0.5cm 6cm 19cm 4cm},clip,page=6]{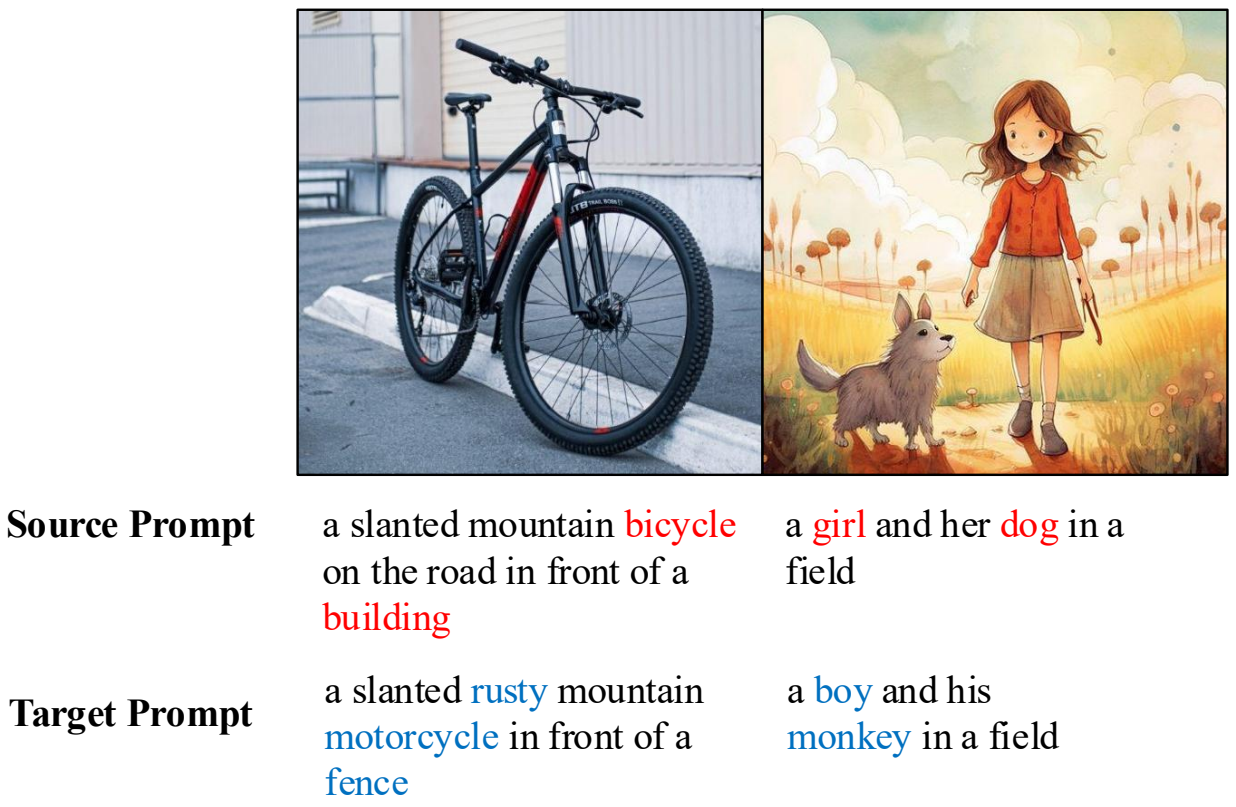}}
    \caption{A sample of \datasetthird{}. The prompts can be leveraged for interleaved co-generation.}
    \label{fig:pie_bench_sample}
\end{figure}

\datasetthird{} was introduced as an image editing benchmark where a source image is given along with a source and target prompts. The goal is to generate corresponding target images that align with both the source image and target prompt (only change what is mentioned in text). We view this dataset as a valuable asset that can be used in a direct application of multimodal generation, interleaved co-generation. Given the high-quality annotations, we can use only the source and target prompts for joint generation. The generated pairs are required to be both correlated and faithful to their text descriptions. A strong multimodal model can be used for interleaved co-generation which facilitates data augmentation and large scale dataset generation for such tasks and benchmarks. A sample of \datasetthird{} is shown in \Cref{fig:pie_bench_sample}. The dataset included 9 different image editing types, e.g., object, colour, and background changes. In total, the benchmark consists of $700$ samples.

\datasetfourth{} is a large-scale medical imaging benchmark consisting of paired radiology images and textual descriptions, commonly used for tasks such as image captioning and visual question answering. Its diversity in anatomical regions, imaging modalities, and clinical conditions also makes it suitable for text-to-image generation. For our experiments, we use a curated subset of the test split, focusing on MRI data from three anatomical groups: brain, joints (hips, shoulders, and elbows), and spine. The final evaluation set consists of 113 brain, 47 joint, and 35 spine MRI scans. All samples were manually verified for quality and relevance.

\subsection{Baselines}
We compare against top-down VAE-based methods (MVAE~\cite{mvae}, mmJSD~\cite{mmjsd}, MoPoE~\cite{elbo}, MMVAE~\cite{mmvae}, MMVAE+~\cite{mmvae_plus}, MMVM VAE~\cite{mmvmvae}, and CMVAE~\cite{vae_diffusion}). For all multimodal VAE baselines, we use the publicly available implementation from~\cite{multivae_package} ($\beta=1$). For diffusion-based baselines, we follow the standard top-down design used in multimodal diffusion models, where all modality streams are coupled within a shared U-Net denoiser and processed in a single reverse diffusion process. A representative example is MM-Diffusion~\cite{mmdiffusion}, for which we report two variants for fair comparison: one based on a Stable-Diffusion-style U-Net matching our backbone, and another using a ResNet-based U-Net. We note that MVAE and mmJSD fail to produce meaningful high-resolution samples on \datasetsecond{}, and are therefore evaluated only on \datasetfirst{}. We also acknowledge the recent work in~\cite{shala}, but do not include quantitative results due to the lack of publicly available code at the time of submission. In our scalability analysis (\Cref{sec:experiments}), we focus on foundation model-based baselines, given the nature of the task and to ensure a fair comparison. We further observe that standard top-down baselines are not directly applicable to text-to-image settings due to their reliance on jointly modeled modalities during training.

\subsection{Evaluation Metrics}
\subsubsection{Coherence}
\label{sec:coherence}
Measuring coherence, i.e., how well two generated representations correspond to one another, is not a direct task. Prior work typically relies on pretrained classifiers to assess cross-modality agreement; if all classifiers predict the same label for each modality, the sample is considered coherent~\cite{elbo,mvae,mld,mmvae,mmvae_plus,limitations_vae,mmjsd,mvrl}. While effective for simple datasets such as \datasetfirst{}, such classifier-based metrics fail in more complex settings like face–mask joint generation. Moreover, previous benchmarks on \datasetsecond{} evaluate coherence only in conditional settings where a ground truth mask exists, using metrics such as the F1 score between generated and true masks~\cite{mld}. These limitations motivate the need for a coherence metric for joint generation on \datasetfirst{}.

\begin{figure}[h]
    \centering
    \resizebox{0.7\columnwidth}{!}{
    \includegraphics[width=\textwidth,trim={3.5cm 6cm 7.5cm 3.5cm},clip,page=1]{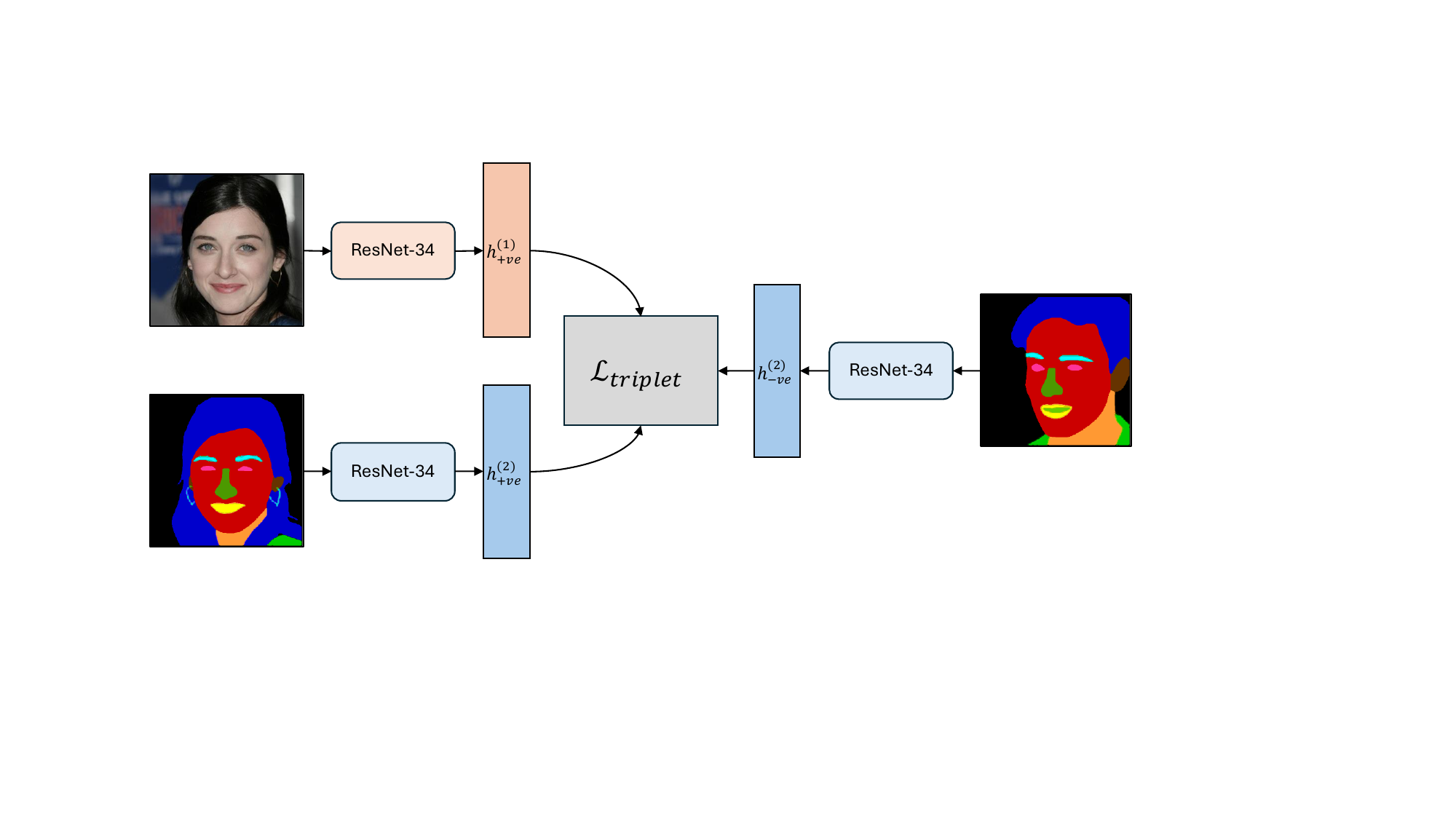}}
    \caption{Contrastive learning framework used to learn the embedding space for coherence measurement. Each representation has its own ResNet-34 model, sharing the architecture not the weights.}
    \label{fig:contrastive}
\end{figure}

Consequently, We develop a learned coherence metric based on a shared embedding space. We train encoders that project the two representations into a joint embedding space, where positive pairs (segmentation masks and faces of the same person) are moved closer, and negative pairs are pushed apart. These encoders are trained using a soft-margin triplet loss of the following form,
\begin{equation}
\label{eq:triplet}
\mathcal{L}_{triplet} = \log\Big(1 + \exp^{\alpha\Big(||h_m^{(1)} - h_m^{(2)}||_2^2 - ||h_m^{(1)} - h_n^{(2)}||_2^2 \Big) } \Big)
\end{equation}
where $\alpha$ is a scaling hyperparameter set to $10$, and $m,n\in\{1,\dots,N\}, \ m\neq n$. Here, $h_m^{(1)}$ and $h_m^{(2)}$ denote the embeddings of the positive mask–face pair, $||\cdot||_2^2$ is the squared Euclidean distance, and ResNet-34 backbones were used to extract $h^{(1)}$ and $h^{(2)}$~\cite{resnet}. A schematic of the contrastive learning setup is shown in \Cref{fig:contrastive}.

After training the ResNet-34 encoders, coherence between any mask–face pair is computed using the distance between their normalized embeddings, where $||h^{(1)}||_2 = ||h^{(2)}||_2 = 1$. The squared Euclidean distance can be simplified to the following,
\begin{align}
\label{eq:euclidean}
\begin{split}
||h^{(1)} - h^{(2)}||_2^2 &= ||h^{(1)}||_2^2 - 2(h^{(1)} \cdot h^{(2)}) + ||h^{(2)}||_2^2 \\
                        &= 2 - 2(h^{(1)} \cdot h^{(2)})
\end{split}
\end{align}
which lies in the range $[0,4]$ due to the $\ell_2$ normalization. To map this distance to a coherence score in $[0,1]$, we compute the following,
\begin{equation}
\label{eq:coherence}
\text{coherence}
= 1 - \frac{||h^{(1)} - h^{(2)}||_2^2}{4}
= 1 - \frac{2 - 2(h^{(1)} \cdot h^{(2)})}{4}
\end{equation}
Thus, higher scores indicate better alignment between the generated representations in the learned space.

\subsubsection{Image Quality Metrics}
We evaluate image fidelity using Fréchet Inception Distance (FID)~\cite{fid}, which measures the distributional similarity between generated and real images using features from a pretrained Inception network. For modalities involving segmentation masks, we additionally report the F1 score to assess segmentation accuracy. In \Cref{sec:scalability}, to evaluate pairwise consistency, we use LPIPS, MSE, and PSNR. These metrics are applied both to measure agreement between generated modality pairs and, in image editing settings (\Cref{sec:scalability}), to compare edited outputs with their corresponding source images. When generating multiple modalities (e.g., 2-4 in \datasetfourth{}), we report the average consistency across all modality pairs. We further assess text-image alignment using CLIP and PubMedCLIP, depending on the domain.

\subsection{Training and Hardware Details}
All models in the main manuscript were implemented in PyTorch~\cite{pytorch}. For \datasetfirst{}, we train a separate DDPM for each modality. We use a U-Net backbone with attention layers at $32\times32$ resolution and train using a standard noise prediction objective with an MSE loss. The forward process follows a DDPM scheduler with 1000 timesteps. Models are trained with AdamW (learning rate $2\times10^{-4}$) and a cosine learning rate schedule with warmup, using a batch size of 128. We maintain an exponential moving average (EMA) of model weights for stable sampling. Training was done for 500 epochs.

For \datasetsecond{}, we used the Adam~\cite{adam} optimizer with $\beta = (0.9, 0.999)$, a learning rate of $1\times10^{-4}$, and a batch size of 32. The face model was a fine-tuned Stable Diffusion v1.5 backbone, while the mask model followed a DDPM-style architecture with five ResNet downsample blocks. The learnable bridge network was implemented as a U-Net that connects the diffusion models of the two modalities. Thus, its input channels depend on the underlying modality (e.g., four channels for the face model due to Stable Diffusion’s latent representation). The U-Net bridge had four downsampling and upsampling layers leading to a depth of 4, with SiLU activations throughout and a base channel of size 32. All training and fine-tuning experiments on \datasetsecond{} were done for 50 epochs, and were conducted on a single AMD MI210 GPU.
%%%%%%%%%%%%%%%%%%%%%%%%%%%%%%%%%%%%%%%%%%%%%%%%%%%%%%%%%%%%%%%%%%%%%%%%%%%%%%
\section{More Qualitative Results}
\label{app:qualitative}
\begin{figure}[H]
    \centering
    \includegraphics[width=0.6\textwidth,trim={0cm 0cm 0cm 4cm},clip,page=1]{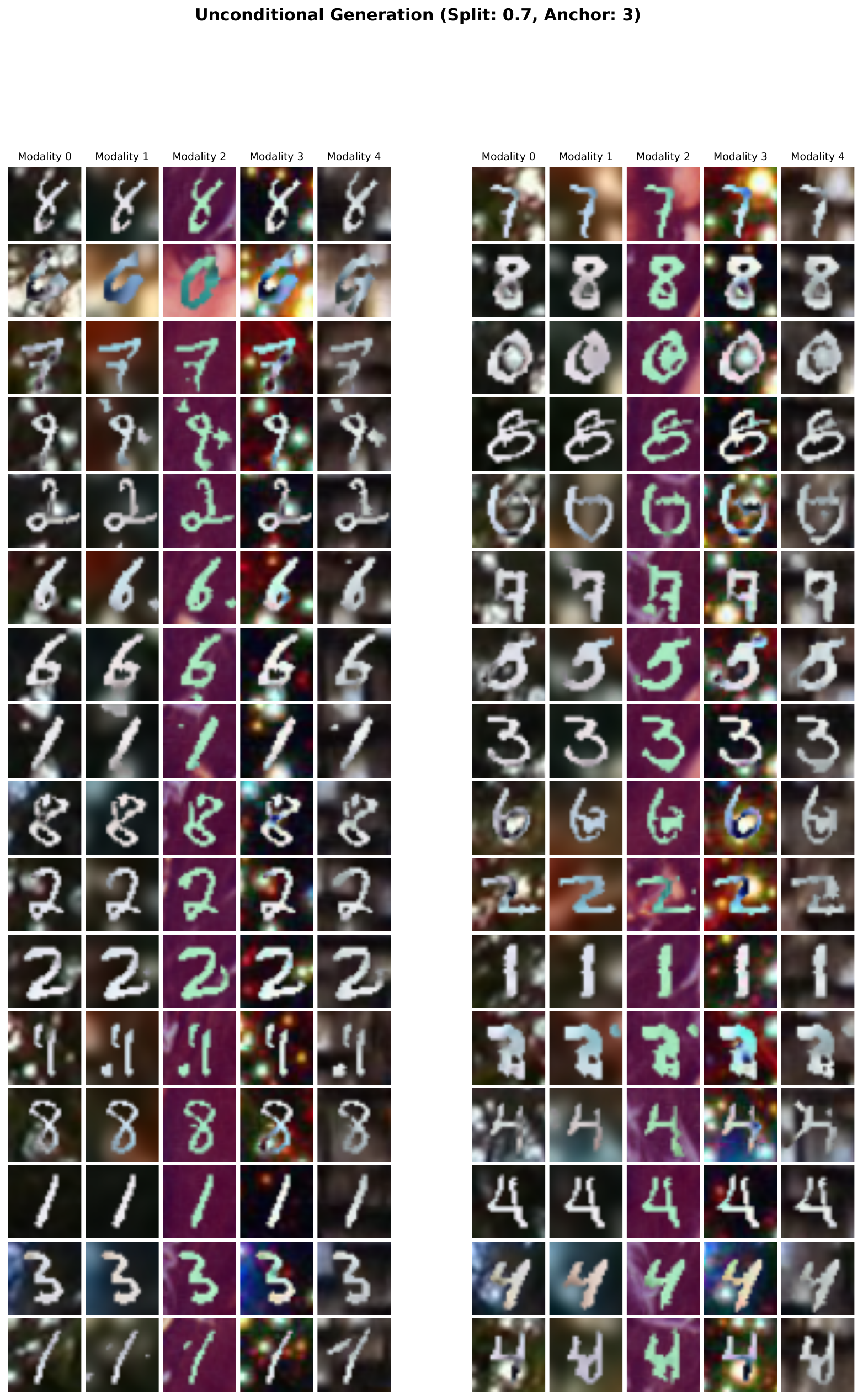}
    \caption{Unconditional generation batched samples on \datasetfirst{}. All modalities share the same structure while only changing the backgrounds.}
    \label{fig:poly_mnits_appendix_vis}
\end{figure}

\begin{figure}[H]
    \centering
    \includegraphics[width=0.95\textwidth,trim={7cm 4cm 6.5cm 4cm},clip,page=6]{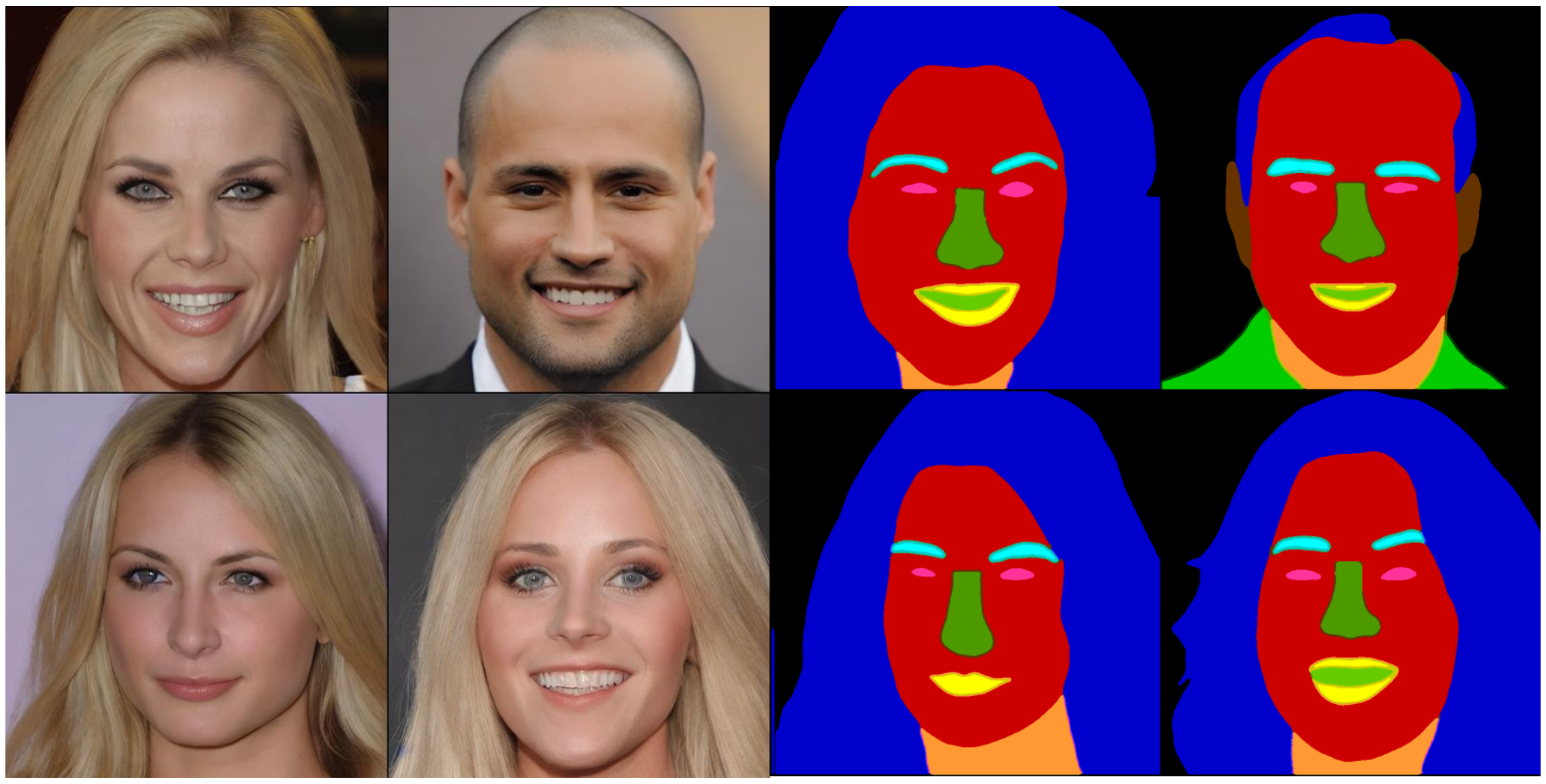}
    \caption{Unlike previous VAE-based approaches, we are able to generate high-resolution $512\times512$ samples. These samples provide much more detail and quality which increase the scalability of joint unconditional generation.}
    \label{fig:zoomed}
\end{figure}

In this section, we present additional qualitative results of \ourmethod{}. More unconditional generations are shown in \Cref{fig:bias,fig:unconditional,fig:poly_mnits_appendix_vis}, where we include randomly selected batches. In \datasetfirst{}, it is clear that the generated digits are the same while only the background changes. Regarding \datasetsecond{}, we observe a slight tendency toward generating blonde female faces, a known bias previously reported in works using the same dataset~\cite{bias1,bias2,bias3}. \ourmethod{} consistently produces coherent face–mask pairs that remain faithful to each other without sacrificing image quality. A zoomed-in example highlighting finer details of the unconditional generation is provided in \Cref{fig:zoomed}. We note these samples were generated with setting $t_{split}=50\%$.

\begin{figure}[H]
    \centering
    \includegraphics[width=\textwidth,trim={1cm 0cm 0cm 0cm},clip,page=7]{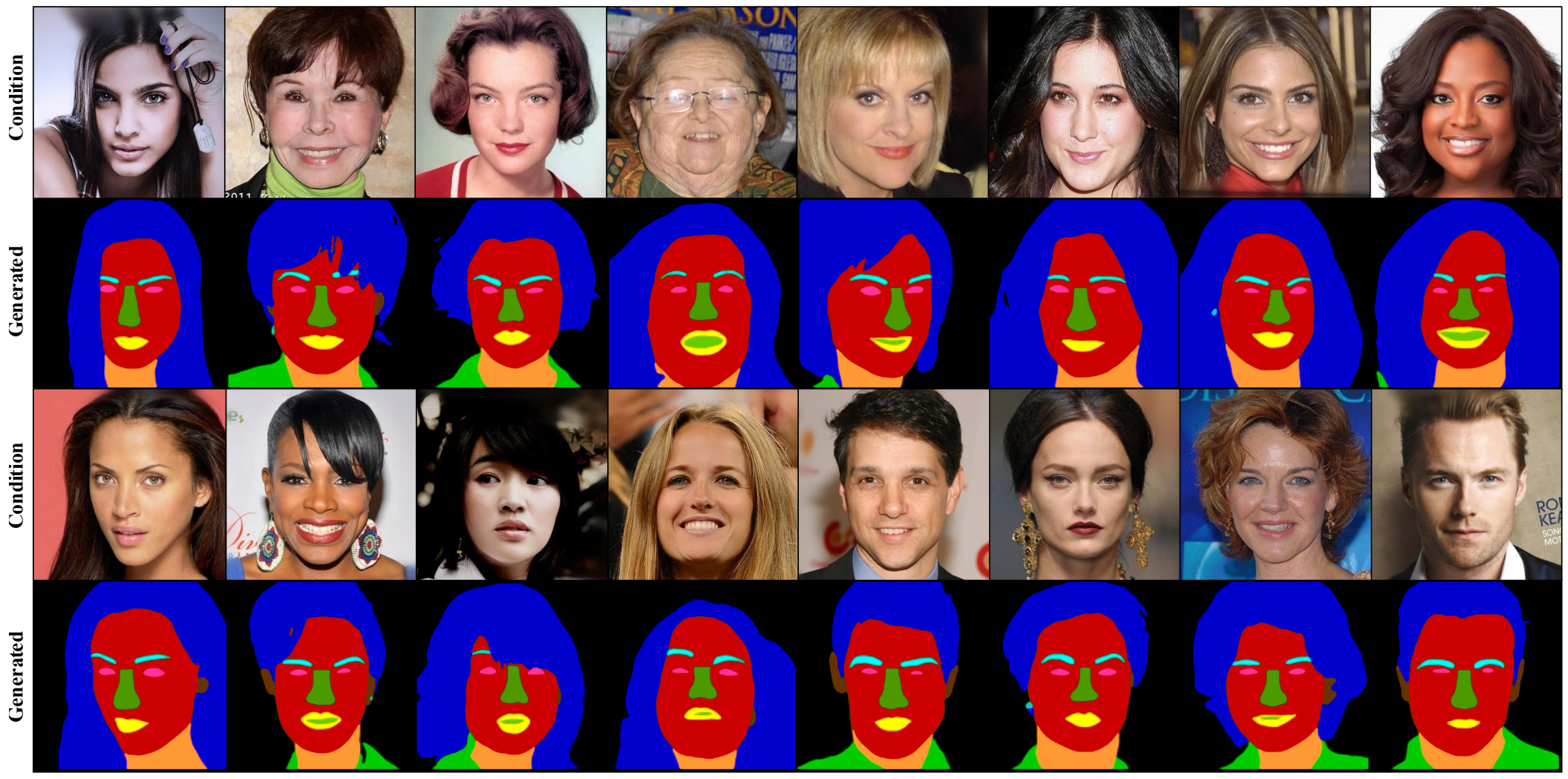}
    \caption{Face $\rightarrow$ mask conditional generation. The generated masks are aligned with the given conditioning faces. The poses are consistent, while some hair details may not be always captured}
    \label{fig:facetomask}
\end{figure}

\begin{figure}[H]
    \centering
    \includegraphics[width=\textwidth,trim={4.5cm 5.5cm 8cm 2.5cm},clip,page=8]{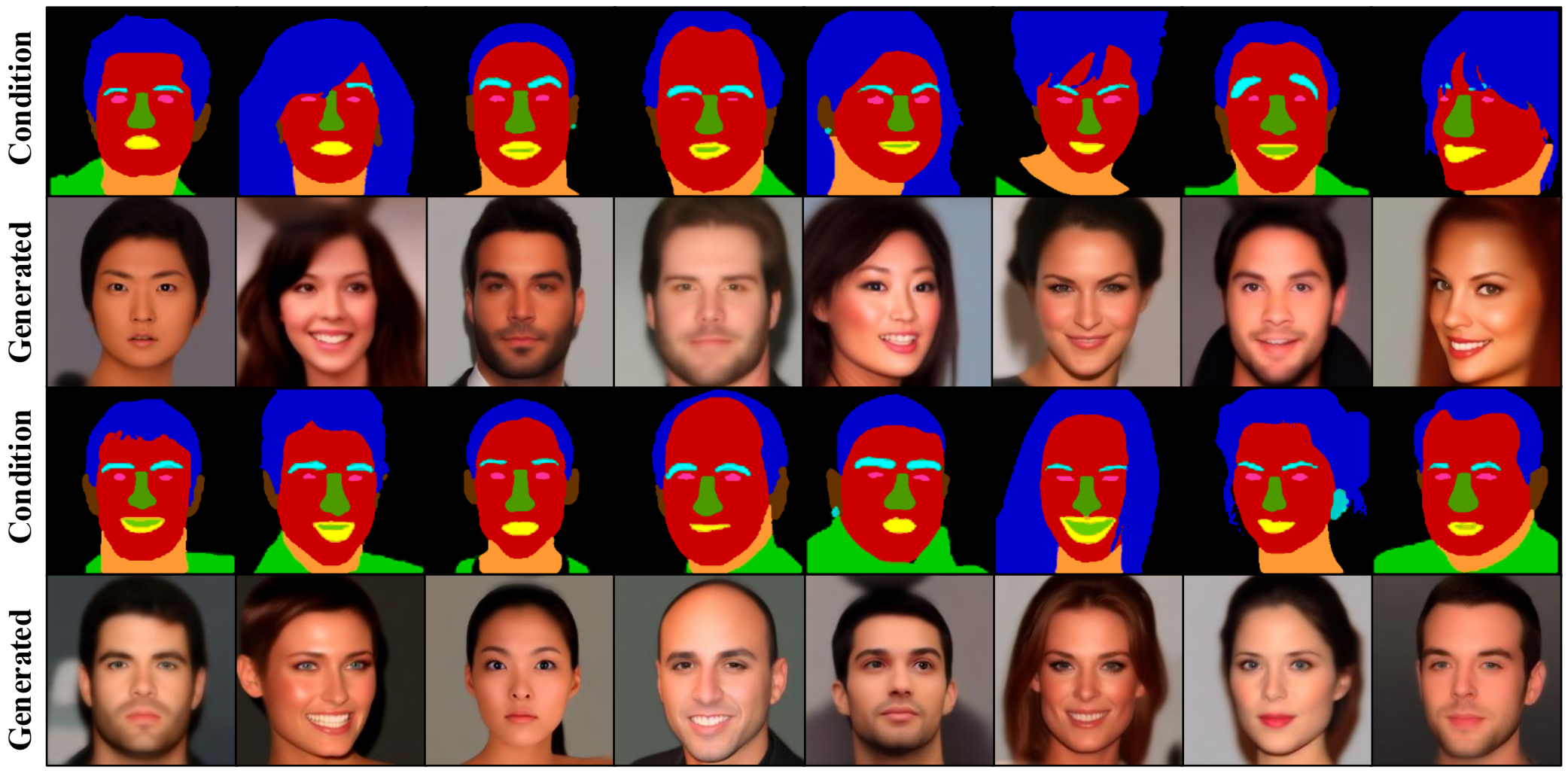}
    \caption{Mask $\rightarrow$ Face conditional generation. The generated faces are aligned with the given conditioning masks. The poses, hair length, and expressions are consistent.}
    \label{fig:masktoface}
\end{figure}

For conditional generation, we visualize results in both directions: mask$\rightarrow$face and face$\rightarrow$mask. \Cref{fig:facetomask} shows the face $\rightarrow$ mask direction, where the first two rows contain the input faces. The generated masks align well with their corresponding faces, particularly in head shape and pose. Conversely, \Cref{fig:masktoface} presents the mask$\rightarrow$face direction. This task is inherently more challenging, as faces contain substantially richer information than segmentation masks. Transporting information from a low-detail modality to a high-detail one requires learning a more complex mapping. Nevertheless, the generated faces in \Cref{fig:masktoface} exhibit strong structural alignment with the input masks and preserve fine facial details.

\subsection{Failure Cases}
\label{sec:failure}
\begin{figure}[H]
    \centering
    \includegraphics[width=\textwidth,trim={3.5cm 10cm 17cm 3cm},clip,page=9]{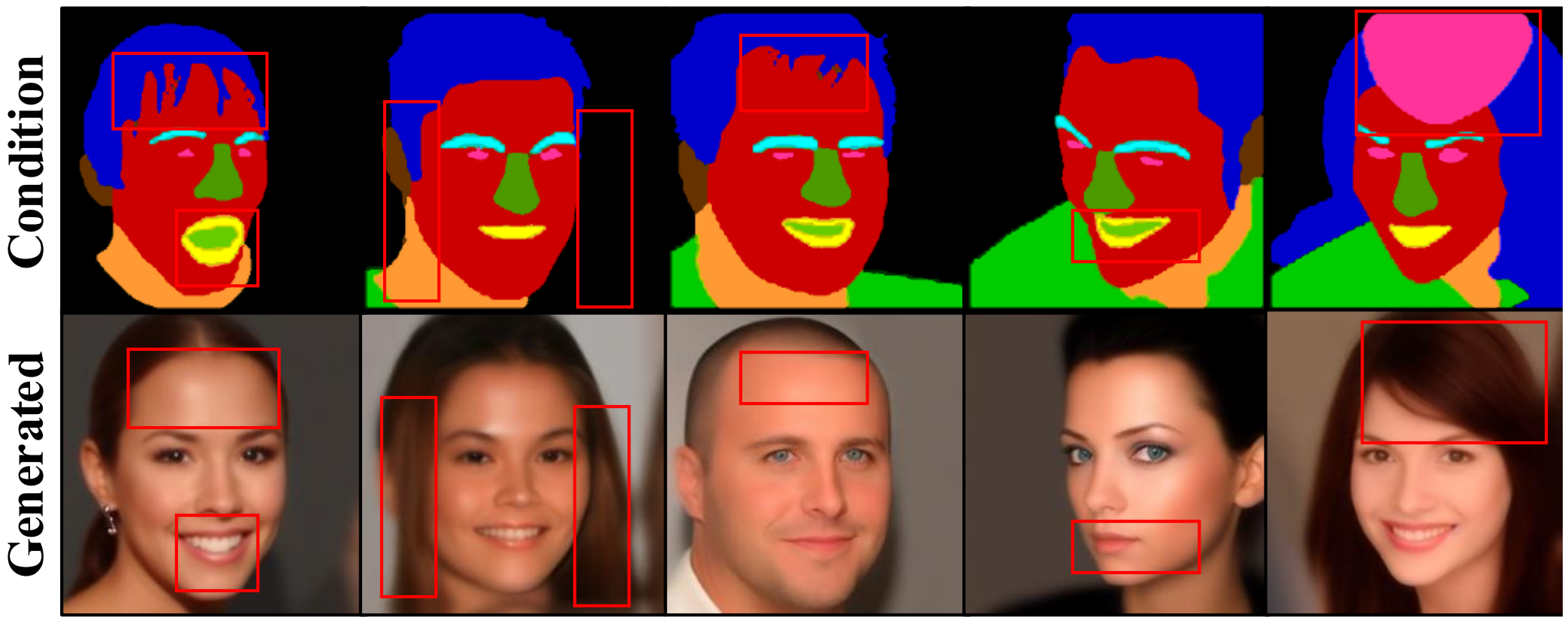}
    \caption{Failure cases in conditional generation on \datasetsecond{}. Some generated faces misalign with the conditioning masks, particularly for accessories such as glasses, hats, and earrings (rightmost image). We also observe occasional inconsistencies in hair styles or lengths.}
    \label{fig:fail}
\end{figure}

We report some failure cases in our generated outputs, particularly in the conditional setting on \datasetsecond{}. As shown in \Cref{fig:fail}, some generated faces deviate from their conditioning masks. For example, exhibiting mismatched hair length, altered facial expressions, or missing accessories. We consistently found that earrings and hats were difficult to reproduce, which we attribute to the inherent difficulty of transporting low-detail segmentation masks to high-detail face images. This limitation suggests that the mask$\rightarrow$face direction may require a stronger or more expressive bridge network.
%%%%%%%%%%%%%%%%%%%%%%%%%%%%%%%%%%%%%%%%%%%%%%%%%%%%%%%%%%%%%%%%%%%%%%%%%%%%%%
\section{Scalability Experiments Details}
\label{app:downstream}
\subsection{Interleaved Co-Generation}
\begin{figure}[H]
    \centering
    \includegraphics[width=\textwidth,trim={0cm 0cm 0cm 5cm},clip,page=1]{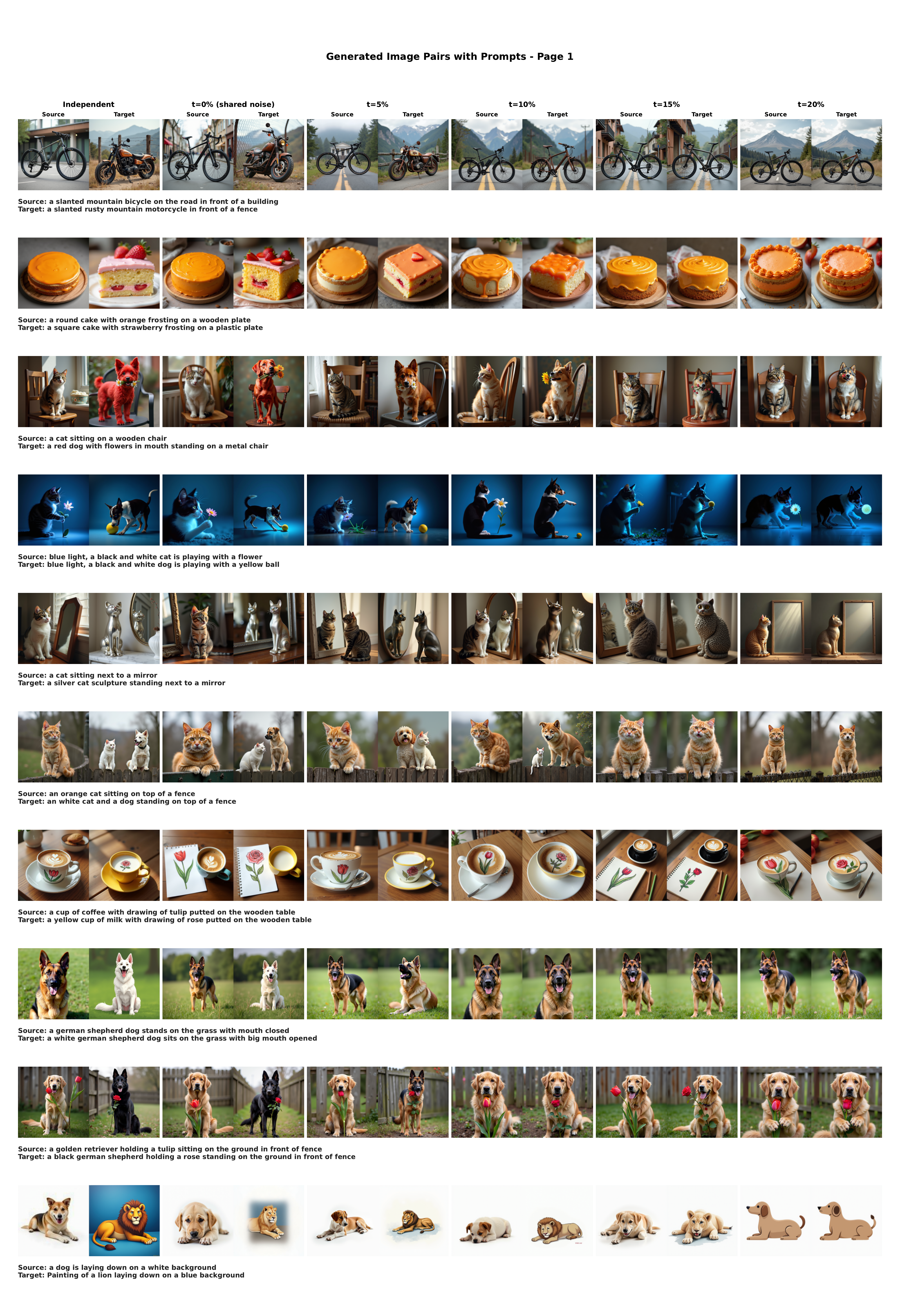}
    \caption{Interleave cogeneration on random samples from \datasetthird{}. Using \ourmethod{} always leads to correlated pairs compared to independently using pretrained models. These results were generated through FLUX.}
    \label{fig:interleave_1}
\end{figure}

For experiments on \datasetthird{}, we use $t_{split}=5\%$ unless stated otherwise. As illustrated in \Cref{fig:interleave_1,fig:interleave_2}, sharing a larger portion of the initial diffusion trajectory leads to more correlated outputs across modalities. We observe that, in large diffusion models, most structural and semantic attributes are established in early timesteps. Consequently, increasing $t_{split}$ improves alignment (e.g., matching pose and orientation), as seen in the later rows of \Cref{fig:interleave_1}. However, this introduces a trade-off: smaller $t_{split}$ values (e.g., $0\%$--$5\%$) better preserve prompt-specific details, while larger values favor stronger cross-modal consistency.
\subsection{Style Coupling and Decoupling}
\label{app:style}
\begin{figure}[H]
    \centering
    \includegraphics[width=0.9\textwidth,trim={4cm 1.5cm 4cm 3cm},clip,page=1]{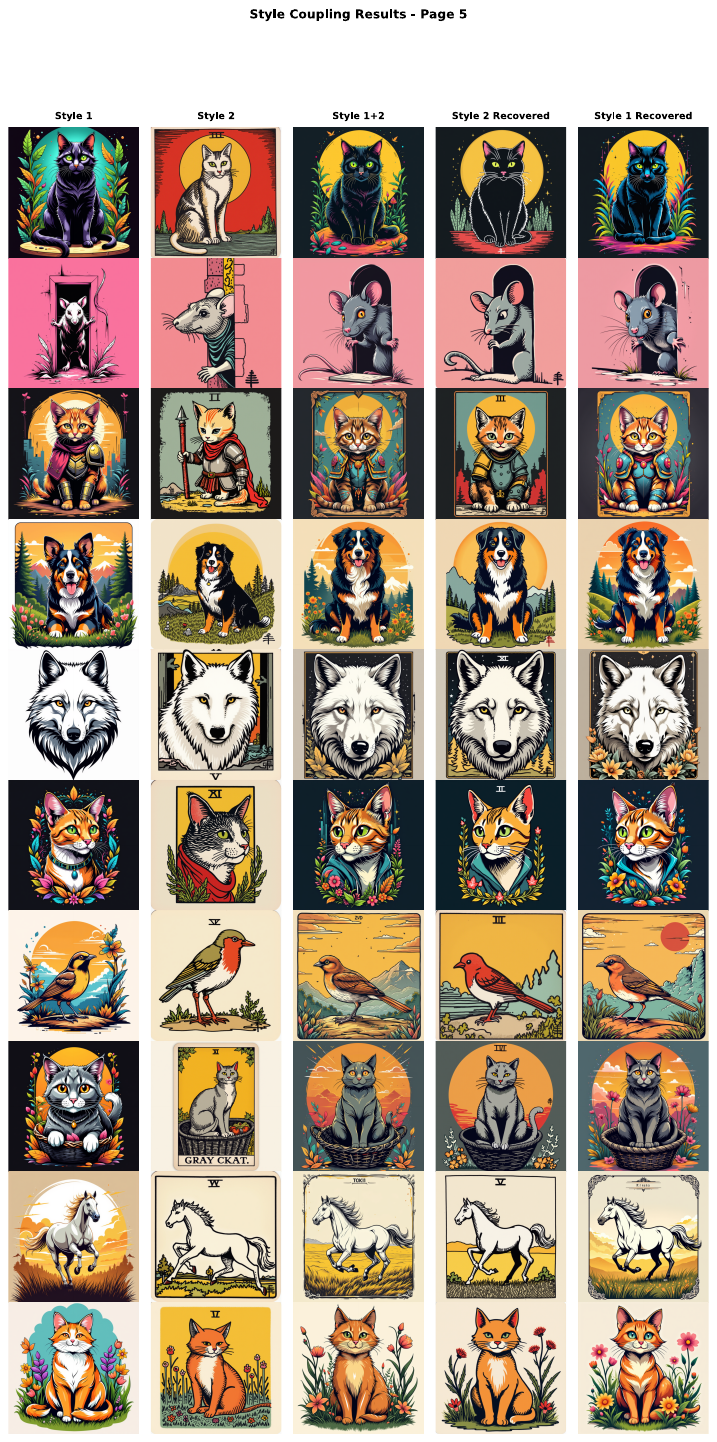}
    \caption{Style coupling and decoupling using parameter-based score interventions with an identity bridge. We sequentially apply two LoRA adaptations at different timesteps, resulting in a composition of both styles (middle). The left columns show individual styles, while the right columns demonstrate decoupling, recovering each style from the combined output.}
    \label{fig:style_1}
\end{figure}

We further demonstrate compositionality through parameter-based score interventions using LoRA. As shown in \Cref{fig:style_1,fig:style_2}, multiple styles can be combined by sequentially applying LoRA adaptations during sampling. Starting from the base model parameters $\mathbf{\theta}_{\text{base}}$, we first inject a LoRA module at timestep $t_{style}^{(1)}$, making it $\mathbf{\theta}_{\text{base}} + \Delta\mathbf{\theta}_1$, followed by a second LoRA intervention at $t_{style}^{(2)}$, leading to $\mathbf{\theta}_{\text{base}} + \Delta\mathbf{\theta}_1 + \Delta\mathbf{\theta}_2$. Intuitively, this process can also be decoupled by adding noise back to a composed sample to an intermediate timestep $t_\text{decouple}$ and continuing sampling with a different style adaptation, enabling partial recovery or modification of individual styles from the combined output. In our experiments, we use $t_{style}^{(1)}=10\%$ and $t_{style}^{(2)}=10\%$. Prompts are taken from the style-change subset of \datasetthird{}.

\subsection{Multimodal MRI Generation}
\begin{figure}[H]
    \centering
    \includegraphics[width=\textwidth,trim={0cm 4cm 13.5cm 0cm},clip,page=8]{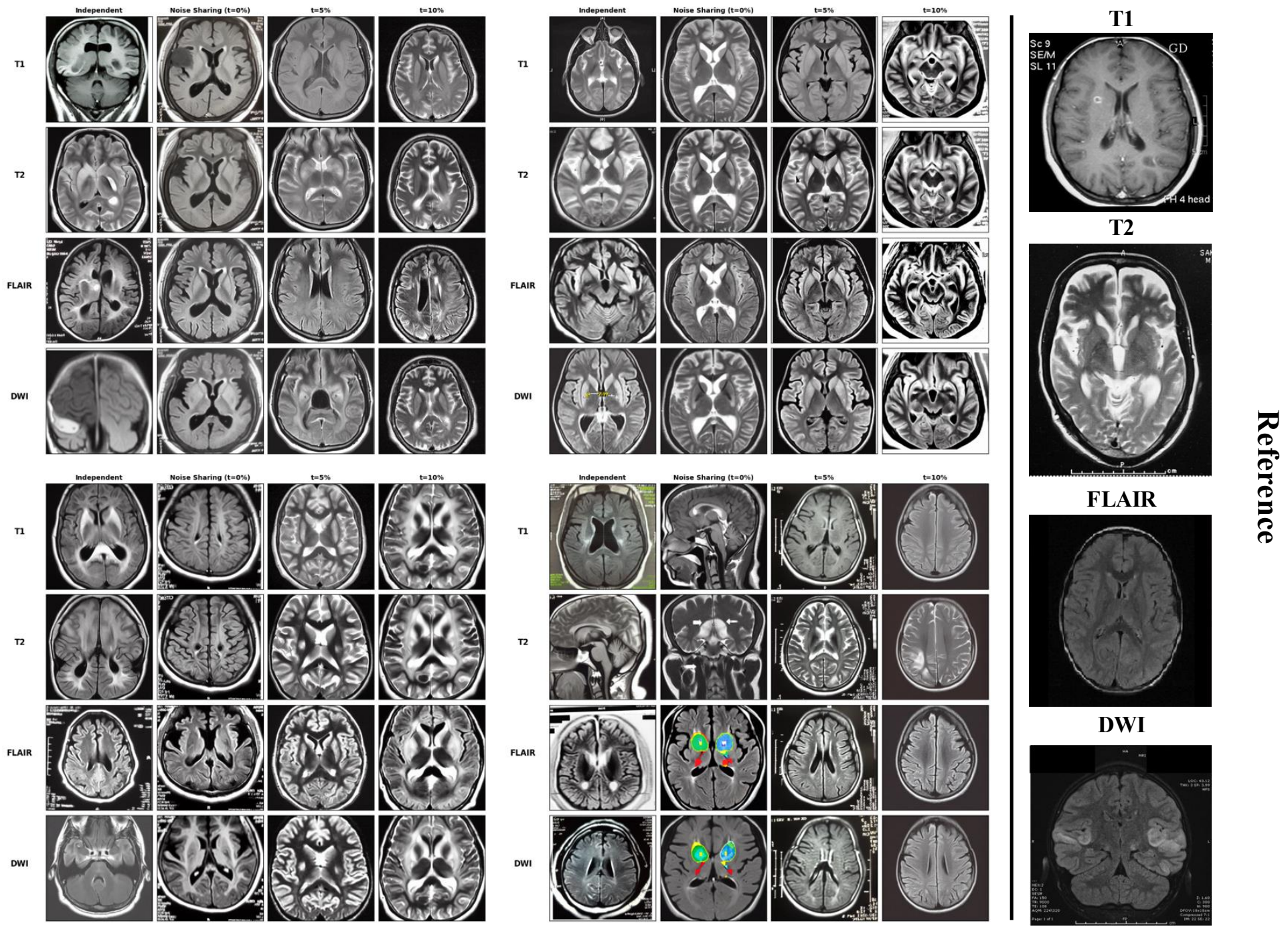}
    \caption{4 Randomly sampled brain MRIs from \ourmethod{}. We provide modality reference images from \datasetfourth{} for those not familiar with MRI contrast modalities. Notice how \ourmethod{} is both faithful to modality specific feature while maintaining coherence.}
    \label{fig:medical}
\end{figure}

Additional qualitative results for multimodal MRI generation on \datasetfourth{} are shown in \Cref{fig:medical}. We include reference samples for each modality for clarity. \ourmethod{} produces consistent anatomical structures across modalities while varying only the contrast (e.g., T1, T2), as observed for both $t_{split}=5\%$ and $10\%$. The quantitative results in \Cref{tab:ROCO_main} correspond to $t_{split}=5\%$, which provides a good balance between coherence and diversity. While these results demonstrate the practical potential of \ourmethod{}, especially in data-scarce domains such as medical imaging, further evaluation with clinically validated metrics and expert assessment is required for real-world deployment.
%%%%%%%%%%%%%%%%%%%%%%%%%%%%%%%%%%%%%%%%%%%%%%%%%%%%%%%%%%%%%%%%%%%%%%%%%%%%%%
\section{Initial Modality Choice}
\label{app:start_point}
\textit{Does the direction of the bridge matter?} How does the initial modality choice before $t_{split}$ influence performance? For example, in face–mask joint generation, should the trajectory originate from the face and then branch to the mask, or should it be the other way around? To answer this, we report the performance of both variants in \Cref{tab:bridge_direction}. We evaluate two settings: one that begins with the face trajectory and splits at $t_{split}=50\%$, and another that starts from the mask and then diverges. As shown, in joint generation, starting from the face produces higher coherence and better FID scores. Although the mask-first variant still functions, its performance lags behind the face-first variant.

We attribute this gap to the inherent domain gap between modalities. Mapping a low-information representation (mask) to a high-information one (face) is fundamentally more challenging than the reverse. This difficulty likely drives the observed performance discrepancy. Therefore, in our main experiments, we adopt the face-first direction as the default bridging strategy (unless in the mask$\rightarrow$face conditional case).
\begin{table}[h]
    \centering
    \caption{Results of multimodal generation on \datasetsecond{} at $t_{split}=50\%$ with different initial trajectories. Starting with the face modality leads to more coherent samples.}
    \label{tab:bridge_direction}
    \resizebox{0.8\columnwidth}{!}{
    \begin{tabular}{@{}l|cc|cc|cc@{}}
    \toprule
    \multirow{2}{*}{\textbf{Initial trajectory}} & \multicolumn{2}{c|}{\textbf{joint}} & \multicolumn{2}{c|}{\textbf{M$\to$F}} & \multicolumn{2}{c}{\textbf{F$\to$M}} \\
    & coherence $\uparrow$ & FID $\downarrow$ & coherence $\uparrow$ & FID $\downarrow$ & coherence $\uparrow$ & F1 $\uparrow$ \\
    \midrule
    \textbf{face-first} & \textbf{0.73} & \textbf{60.84} & - & - & \textbf{0.72}  & \textbf{0.85} \\
    \textbf{mask-first} & 0.61 & 97.20 & \textbf{0.60} & \textbf{107.22} & -  & - \\
    \bottomrule
    \end{tabular}}
\end{table}
%%%%%%%%%%%%%%%%%%%%%%%%%%%%%%%%%%%%%%%%%%%%%%%%%%%%%%%%%%%%%%%%%%%%%%%%%%%%%%
\section{Societal Impact}
\label{app:impact}
Recent advances in generative modeling have led to widespread adoption across many applications. Our framework has the potential to support positive downstream uses, such as controllable content creation, personalized user experiences, and new artistic workflows, as demonstrated by our experiments in \Cref{sec:scalability}. These capabilities highlight clear opportunities for creative and valuable scientific tools built on multimodal generation. However, as with any model capable of manipulating or synthesizing faces, there is a risk of misuse in harmful contexts, including unauthorized identity manipulation or deceptive content creation. We emphasize the importance of responsible deployment, transparency about synthetic outputs, and adherence to ethical guidelines when applying such models.
%%%%%%%%%%%%%%%%%%%%%%%%%%%%%%%%%%%%%%%%%%%%%%%%%%%%%%%%%%%%%%%%%%%%%%%

\begin{figure}[H]
    \centering
    \includegraphics[width=\textwidth,trim={0cm 0cm 15cm 0cm},clip,page=3]{figures/figs.pdf}
    \caption{Unconditional generation batched samples. In all cases, the poses and face structure is shared between the segmentation masks and faces. We notice a slight bias towards blonde females in unconditional generation}
    \label{fig:bias}
\end{figure}

\begin{figure}[H]
    \centering
    \includegraphics[width=\textwidth,trim={0cm 0cm 15cm 0cm},clip,page=4]{figures/figs.pdf}
    \caption{Unconditional generation batched samples. In all cases, the poses and face structure is shared between the segmentation masks and faces. We notice a slight bias towards blonde females in unconditional generation}
    \label{fig:unconditional}
\end{figure}

\begin{figure}[H]
    \centering
    \includegraphics[width=\textwidth,trim={0cm 0cm 0cm 5cm},clip,page=2]{figures/interleave.pdf}
    \caption{Interleave cogeneration on random samples from \datasetthird{}. Using \ourmethod{} always leads to correlated pairs compared to independently using pretrained models. These results were generated through FLUX.}
    \label{fig:interleave_2}
\end{figure}

\begin{figure}[H]
    \centering
    \includegraphics[width=0.9\textwidth,trim={4cm 1.5cm 4cm 3cm},clip,page=2]{figures/style.pdf}
    \caption{Style coupling and decoupling using parameter-based score interventions with an identity bridge. We sequentially apply two LoRA adaptations at different timesteps, resulting in a composition of both styles (middle). The left columns show individual styles, while the right columns demonstrate decoupling, recovering each style from the combined output.}
    \label{fig:style_2}
\end{figure}

%\newpage
%\input{checklist.tex}

\end{document}